\RequirePackage{fix-cm}
\documentclass[twocolumn,compsoc]{cvm}

\graphicspath{{figures/},{figures/cvm/}}

\def\ManuscriptInfo{}

\def\PaperTitle{G2MF-WA: Geometric Multi-Model Fitting with Weakly Annotated Data}

\def\AuthorNames{Chao Zhang$^1$ (\rm{\Letter}), Xuequan Lu$^2$, Katsuya Hotta$^1$, and Xi Yang$^3$}

\def\AuthorAdress{$1\quad $ & University of Fukui, Fukui,
    910-8507, Japan. E-mail: zhang@u-fukui.ac.jp (\rm{\Letter}), k-hotta@monju.fuis.u-fukui.ac.jp.\\
    $2\quad $ & Deakin University, Waurn Ponds, 3216, Australia.
    E-mail: xuequan.lu@deakin.edu.au.\\
    $3\quad $ & The University of Tokyo, Tokyo, 113-8656, Japan.
    E-mail: earthyangxi@gmail.com.\\
}

\def\firstheader{\hfill \\}

\headevenname{{C. Zhang, X. Lu, K. Hotta, X. Yang \etal} }

\usepackage{times}
\usepackage{epsfig}
\usepackage{graphicx}
\usepackage{amssymb}
\usepackage{amsmath}
\usepackage{newtxtext}
\usepackage[varg]{newtxmath}
\usepackage{mathtools}
\usepackage{amsfonts}
\usepackage{algorithm}
\usepackage{algorithmic}
\usepackage{commath}
\usepackage{dsfont}
\usepackage{xcolor,colortbl}

\renewcommand{\algorithmiccomment}[1]{\bgroup\hfill//~#1\egroup}

\definecolor{t3}{RGB}{230,255,255}
\definecolor{t2}{RGB}{150,255,255}
\definecolor{t1}{RGB}{100,255,255}
\definecolor{e3}{RGB}{255,230,255}
\definecolor{e2}{RGB}{255,150,255}
\definecolor{e1}{RGB}{255,100,255}

\newcommand{\cz}[1]{{\textcolor{black}{#1}}}
\newcommand{\czcz}[1]{{\textcolor{black}{#1}}}

\usepackage{comment}

\begin{document}
\MakePageStyle
\MakeAbstract{In this paper we attempt to address the problem of geometric multi-model fitting with resorting to a few weakly annotated (WA) data points, which has been sparsely studied so far. In weak annotating, most of the manual annotations are supposed to be correct yet inevitably mixed with incorrect ones. The WA data can be naturally obtained in an interactive way for specific tasks, for example, in the case of homography estimation, one can easily annotate points on the same plane/object with a single label by observing the image. Motivated by this, we propose a novel method to make full use of the WA data to boost the multi-model fitting performance. Specifically, a graph for model proposal sampling is first constructed using the WA data, given the prior that the WA data annotated with the same weak label has a high probability of being assigned to the same model. By incorporating this prior knowledge into the calculation of edge probabilities, vertices (i.e., data points) lie on/near the latent model are likely to connect together and further form a subset/cluster for effective proposals generation. With the proposals generated, the $\alpha$-expansion is adopted for labeling, and our method in return updates the proposals. This works in an iterative way. Extensive experiments validate our method and show that the proposed method produces noticeably better results than state-of-the-art techniques in most cases.}

\MakeKeywords{geometric multi-model fitting; weak annotation; multi-homography detection; two-view motion segmentation}
\section{Introduction}
Geometric model fitting, aims to fit a model with data which contains both inliers and outliers. A well-known example is RANSAC \cite{Fischler1981RandomCartography}, the main idea of which is to generate a number of random model proposals and select the best solution which holds the largest inlier set based on an inlier threshold. The geometric multi-model fitting task further 
assumes that multiple models are embedded in the input data. Multi-model fitting algorithms have to optimize the global solution, rather than taking a greedy strategy to maximize the inliers for single model exploration like RANSAC. To evaluate the numerous possible solutions, one common way is to design an energy function \cite{boykov1999fast,delong2012fast,Isack2012,amayo2018geometric}, such that an approximate solution can be achieved via energy minimization/maximization by balancing the geometric errors (data fidelity) and the regularity of inlier clusters (e.g., smoothness, complexity). Although finding the optimal solution is NP-hard \cite{Isack2012}, $\alpha$-expansion \cite{boykov1999fast} provides a powerful alternative to find solutions with guaranteed approximation bounds over a given set of model proposals. However, the quality of the solution and the convergence largely depend on the quality of the proposals which 
greatly influences the overall efficiency and effectiveness. 

\begin{figure}[t]
	\centering
	\includegraphics[width=1\linewidth]{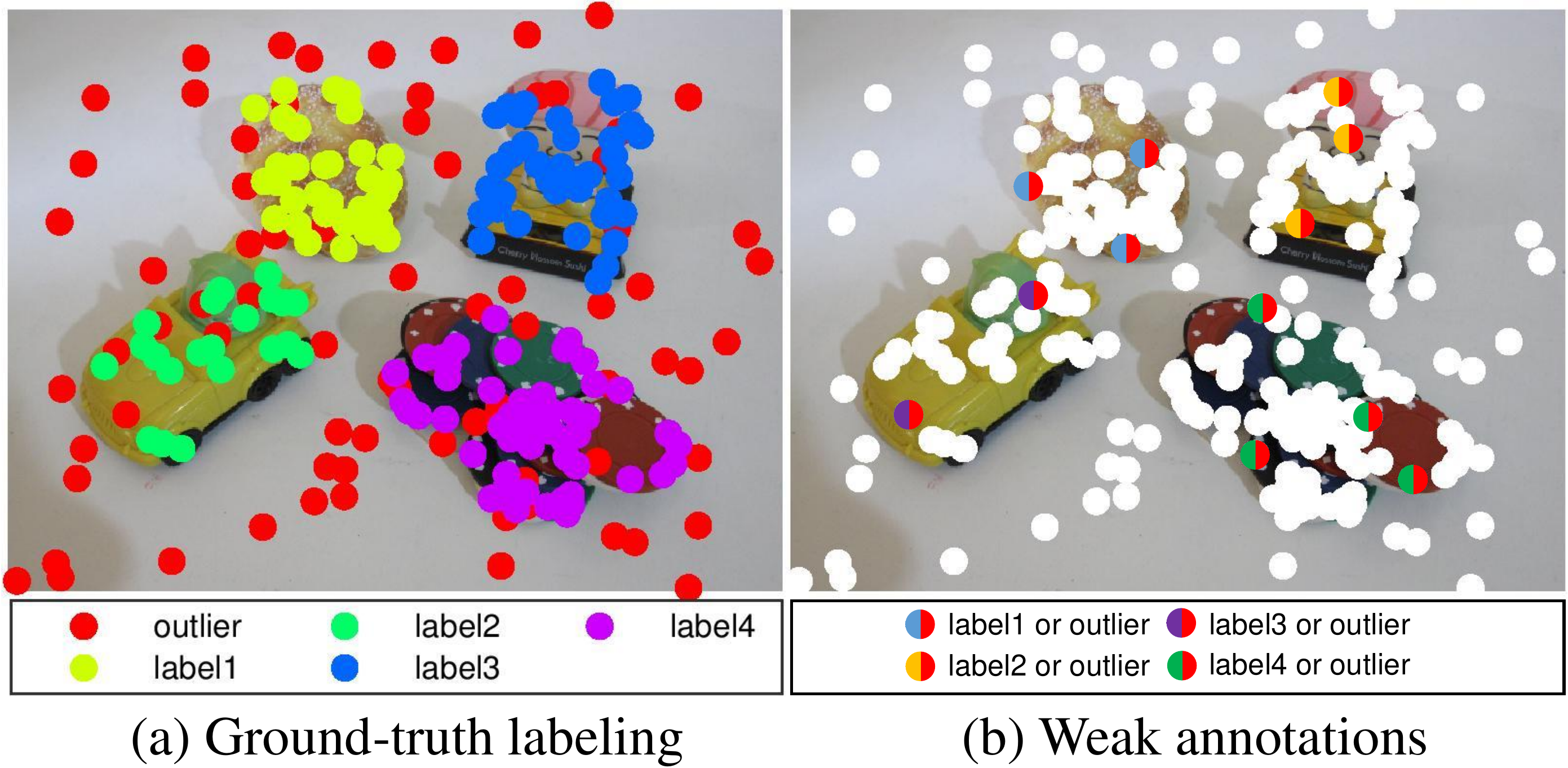}
	\caption{Example of weak annotations in the two-view motion segmentation task (only one view is shown here). (a) Ground-truth labeling. Four objects are with \cz{independent} motions, and the movement of the camera induces the outliers. (b) \cz{Detected feature points which are represented by circles including weakly annotated ones (bi-colored) and unlabeled ones (white).} The annotator annotates points on an object with a corresponding \cz{bi-colored weak label, based on the fact that they can not distinguish them from outliers}. Note that the number of weak labels in (b) is not necessarily equal to the number of ground-truth labels in (a). Best viewed in color. }
	\label{fig:intro}
\end{figure}

Most of the methods attempted to improve the quality of model proposals by sampling ``clean'' subsets of data points from the input data. This paper, however, claims that weakly annotated data (WA data) which has been sparsely treated so far in multi-model fitting tasks, can be exploited to improve the quality of the proposals and further the fitting performance. In Fig. \ref{fig:intro}, we show an example of weak annotations in the scenario of motion segmentation. One can observe that the four objects in the image are with independent motions according to a pair of two-view images (i.e., two-view motion segmentation problem, only one view is shown here). Due to the camera shake, the movement of feature points on each object may also involve camera motion, yielding the inliers biased and outlier points on objects are hardly distinguishable from inliers by observation. Nevertheless, the annotator can at least tell that points on the same object either belong to the outlier model or a shared motion model. We refer to such inaccurate annotations as weak annotations (WA) in this work (Fig. \ref{fig:intro} (b)). To take advantages of the WA data, two priors are observed and exploited: (1) data points with the same weak label has a high probability of belonging to the same model. (2) Data points with different weak labels have a low probability of belonging to the same model, except for the outlier model.

As the main technical contribution of this paper, we \cz{propose to independently construct a proposal sampling graph} with only WA data, apart from the adjacent graph of the $\alpha$-expansion \cite{boykov1999fast}. \cz{Inspired by the random clustering model \cite{pham2014random}, the sampling graph probabilistically forms subsets/clusters to generate model proposals controlled by edge probabilities}. By incorporating the prior assumptions mentioned above into the update procedure of the edge probabilities, proposals with high quality can be generated, therefore leading to appealing fitting performance. Extensive experiments validate our approach, and show that it mostly outperforms state-of-the-art methods, in terms of accuracy and runtime.

\section{Related Work}

In this section, we first review two popular categories of techniques for multi-model fitting in Sec. \ref{sec:greedy} and \ref{sec:energy}, respectively. Then, we focus on investigating efficient proposal generators, which are important to both of the above fitting techniques and closely related to our study.
\subsection{Greedy Methods}
\label{sec:greedy}
RANSAC \cite{Fischler1981RandomCartography} and its variants \cite{chum2005matching,nister2005preemptive,brachmann2017dsac} belong to a category which aims to estimate the parameters of one model with the greedily largest number of inliers (i.e., maximum consensus). The main philosophy is to iterate the following two steps: (1) generating ``good'' proposals based on proposal-verification, (2) refining the proposals by maximum consensus. Since RANSAC is efficient in the case of fitting a single model, many researchers work on extending it to multi-model cases \cite{torr1998geometric,vincent2001detecting,zuliani2005multiransac}. In \cite{torr1998geometric,vincent2001detecting}, the standard RANSAC is ran sequentially and the model with maximum consensus in the current round will be removed in the next round. On the other hand, in \cite{zuliani2005multiransac}, the authors claimed that a parallel fashion is more stable than the sequential fashion in dealing with multiple models. Other common greedy methods are \cite{toldo2008robust,magri2014t,magri2016multiple}. The authors of \cite{magri2016multiple} tried to solve the multi-model fitting problem in terms of set coverage. In \cite{toldo2008robust,magri2014t}, data points lay on/near the same model are considered to share similar preferences (a vector consists of proposals sorted according to the residual). This is an important property for grouping points into the same model, which has also been taken into account for edge probability calculation in our work. 
\subsection{Energy-based Methods}
\label{sec:energy}
\czcz{It has become predominant} to solve the fitting problem under optimization frameworks in recent years. Energy-minimization based methods \cite{yu2011global,Isack2012,delong2012fast,amayo2018geometric} design a global energy function (i.e., the objective function) to evaluate solutions, and the optimal solution is supposed to be found with the minimum energy value. The energy function can be composed of different terms such as the data fidelity term \cite{Isack2012}, smoothness term \cite{Isack2012}, and label term \cite{delong2012fast}. In \cite{yu2011global}, the multi-model fitting of geometric structures is formulated by the quadratic program, in which the data fidelity and the similarity between associated data are balanced. Most of the energy-based methods follow a two-stage strategy: (1) generating a large number of proposals with random subsets of data, (2) evaluating the quality of the proposal by certain likelihood functions \cite{jian2007two}. The proposals
with large likelihood values are sampled and used for labeling. For more multi-model fitting methods, interested readers can refer to a survey paper \cite{nieuwenhuis2013survey}.
\subsection{Proposal Generation}
\label{sec:proposal}
Either of the above categories of methods demands high-quality proposals to decrease fitting error or increase the convergence speed. In particular, in the case of a large data set, it is computationally impractical to exhaustively evaluate each possible model proposal with full data. Importantly, the number of ground-truth models is usually unknown in real-world tasks. Such challenges motivate the fitting algorithms to discretize the sampling space using subsets of the data and generate proposals by fitting with each subset. Although the proposals can be updated iteratively in a propose-and-refine fashion, different initialization of proposals can lead to differing convergence results. The generation of high-quality proposals in the early stage, an important problem in computer vision from a general perspective of robust fitting \cite{meer2004robust}, is crucial for improving the final labeling results for both greedy and energy-based methods.

Instead of full random initialization, many works improve the quality of proposals by utilizing information from inliers \cite{chum2003locally}, certain meta-information (e.g., keypoint matching score) \cite{tordoff2005guided,chum2005matching,chin2010accelerated} or sparsity prior \cite{figueiredo2002unsupervised}. The main factors that affect the quality of proposals can be: (1) the inlier rate of a subset, and (2) the size of a subset. In other words, a large subset with a high inlier rate can lead to a high-quality proposal. The contradiction here is that minimal subsets with high inlier rates may amplify the noise \cite{meer2004robust} while a large subset with a low inlier rate will decrease the efficiency of proposal sampling \cite{meer2004robust} and lead to an exponential growth of computational cost. Pham et al. \cite{pham2014random} alleviated the above contradiction by using the Swendsen-Wang method \cite{swendsen1987nonuniversal} to improve the efficiency of proposal sampling.

In this paper, we propose to generate high-quality proposals with WA data, which is from a new perspective. Note that a proposal generated from an outlier-free sample is not guaranteed to be consistent with all inliers in practice, which makes the problem challenging even with the aid of weakly annotated data. We will elaborate how to elegantly handle it in next section.

\section{Our Approach}
To produce decent multi-model fitting results, our motivation is to effectively generate subsets for generating model proposals with high inlier rates under the guidance of partially and inaccurately labeled data (i.e., weak annotations). 
The multi-model fitting problem can be formulated mathematically. Specifically, given input data set $\mathcal{X}=\{x_i\}_{i=1}^{N}$, which contains outliers and weak annotated data (WA data) $\hat{\mathcal{X}}=\{\hat{x_u}\}_{u=1}^{Z}$, multiple unknown models $\mathcal{M}=\{m_k\}_{k=1}^{K}$ are embedded and need to be estimated ($m_1$ is the outlier model, $K$ is also unknown). Each $x_i$ is assigned to a certain $m_k$ and the assignment procedure is referred to as labeling, with its result denoted by $\mathcal{L}=\{l_i\}_{i=1}^{N}$. Each $l_i$ indicates that $x_i$ is assigned to a certain model in $\mathcal{M}$. From the perspective of energy minimization, this can be generally solved by minimizing the following global energy function. 
\begin{equation}
\label{eq:EM}
E(\mathcal{X},\mathcal{M},\mathcal{L}) = 
\underbrace{D(\mathcal{X},\mathcal{M},\mathcal{L})}_\text{\textrm{data fidelity}} + 
\underbrace{S(\mathcal{X},\mathcal{L})}_\text{\textrm{smoothness}}+
\underbrace{O(\mathcal{M})}_\text{\textrm{complexity}},
\end{equation}
where the data term $D$ is usually a distance or error metric to evaluate the data fidelity according to the labeling result. In this paper, the residuals in the form of Sampson distance \cite{hartley2003multiple} are used. A larger $D$ indicates a larger error of assigned labels. The smoothness term $S$ is based on the prior assumption that spatially close neighbors are assumed to have the same label with a high probability. The neighbors are defined by a neighborhood system (e.g., Delaunay triangulation), with weights on edges indicating how likely two data points are from the same model. A larger $S$ indicates worse local smoothness. The complexity term $O$ penalizes the complexity (e.g., number of models) of the whole optimization task. The solution exploration by minimizing $E$ is effective and has been validated in many works \cite{Isack2012,pham2014random}. We aim to explore the solution more effectively and efficiently with the help of the WA data, which can be achieved interactively with manual operations, based on the natural fact that the feature points belong to different models are mostly visually distinguishable (e.g., points on images belong to different objects, structures, etc).  

\begin{figure}[t]
	\centering
	\includegraphics[width=1\linewidth]{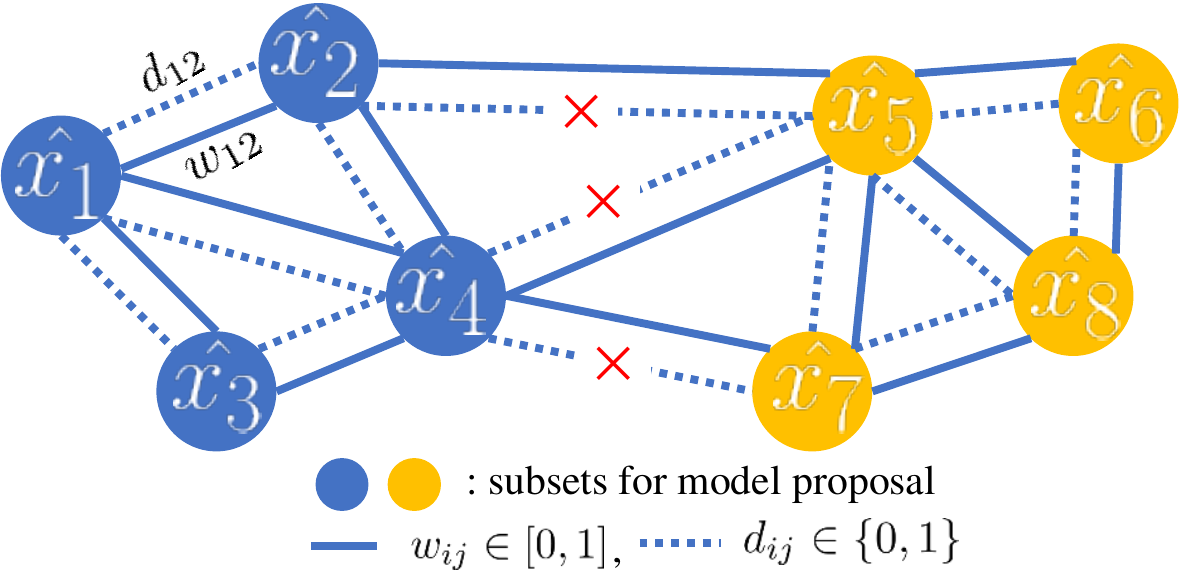}
	\caption{The proposal sampling graph. WA data points $\hat{x_i}$ are the vertices and are divided into two independent subsets (blue or yellow here), according to the connectivity of dotted edges $d_{ij}$. $\{d_{ij}\}$ are the binary \cz{``bonds''} for the random cluster model \cite{pham2014random}. Here, $d_{25}, d_{45}, d_{47}$ are all equivalent to zero and others are one. $w_{ij}$ denotes the edge probability between the $i$-th and $j$-th vertices \cz{to probabilistically determine the values of $\{d_{ij}\}$}. Clusters introduced by $\{d_{ij}\}$ form model proposals. }
	\label{fig:anno}
\end{figure}

\subsection{Proposal Sampling Graph with WA Data}
\label{sec:psg}
The solution quality of $min_{\mathcal{M},\mathcal{L}}E(\mathcal{X},\mathcal{M},\mathcal{L})$ is closely related to the quality of proposals generated by the data subsets sampled from $\mathcal{X}$. We build a sampling Graph $\mathcal{\hat{G}}=(v,e)$ from $\mathcal{\hat{X}}$, apart from the adjacency Graph $\mathcal{G}$ built from $\mathcal{X}$. For clarity, we illustrate $\mathcal{\hat{G}}=(v,e)$ in Fig. \ref{fig:anno} under a specific neighbor system. In our implementation, Delaunay triangulation is adopted for constructing the neighbor system as suggested by \cite{delong2012fast}. $d_{ij}$ can be treated as a ``switch'' to turn on and off the connection between vertices, with the probability determined by corresponding $w_{ij}$. A certain sample of $\{d_{ij}\}$ links to a clustering result of $\mathcal{\hat{X}}$, and each subset is used to calculate the model proposal $\theta_g$ depending on the task setting. For example, in the case of multi-homography detection task, the homography proposal can be solved by the direct linear transformation (DLT) method \cite{hartley2003multiple} as long as the number of points in a subspace is four or above. $w_{ij}$ objectively indicates how likely a pair of points $(\hat{x_i}, \hat{x_j})$ belong to a same model. In unsupervised situation, a common idea is to assume the preferences of inliers from the same model over a set of so-far-generated proposals are correlated \cite{pham2014random,chin2010accelerated}. Specifically, let $\mathcal{H}=\{\theta_g\}_{i=g}^{G}$ be the set of 
the generated proposals during the iteration, and the residuals of $\hat{x_u}\in \mathcal{\hat{X}}$ with respect to each proposal $\theta$ in $\mathcal{H}$ form a vector
\begin{equation}
\label{eq:residuals}
r^{\hat{x_u}}=(r_{1}^{\hat{x_u}},r_{2}^{\hat{x_u}},\cdots,r_{G}^{\hat{x_u}}).
\end{equation}

This can be viewed as a preference vector quantified by residuals. By sorting $r_i^{\hat{x_u}}$ in an ascending order and leave out the elements after the \mbox{$h$-th} place, the preference permutation can be represented as 
\begin{equation}
\label{eq:permutation}
p^{\hat{x_u}}=(p_{1}^{\hat{x_u}},p_{2}^{\hat{x_u}},\cdots,p_{h}^{\hat{x_u}}),
\end{equation}
where each element in $p^{\hat{x_u}}$ is a proposal in $\mathcal{H}$. Then $w_{ij}$ can be updated by the correlation \cite{pham2014random} between $p^{\hat{x_u}}$ and $p^{\hat{x_v}}$ in an online fashion as, 
\begin{equation}
\label{eq:Auv}
w_{ij}=\vert p^{\hat{x_u}} \cap p^{\hat{x_v}} \vert/h.
\end{equation}

The main drawback of Eq. \ref{eq:Auv} is that the confidence of $w_{ij}$ grows with the increase of $G$. At the beginning of any iterative algorithms, $w_{ij}$ can be with low confidence which hinders the whole algorithm from converging to the correct solution. We propose to utilize the prior knowledge brought by the WA data to make $w_{ij}$ more confident. That is, WA data with a same weak label has a high probability to be assigned to the same model and vice versa. By incorporating this property, Eq. \ref{eq:Auv} can be reformulated as a weighted function, 
\begin{equation}
\label{eq:ourW}
w_{ij}=\lambda\vert p^{\hat{x_i}} \cap p^{\hat{x_j}} \vert/h+(1-\lambda)Pr(\hat{x_i},\hat{x_j}),
\end{equation}
where
\begin{equation}
Pr(\hat{x_i},\hat{x_j})=\begin{cases}
1-\sigma \ \ \ \hat{x_i} \mathrm{,} \hat{x_j} \mathrm{\ are \ with \ the \ same \ weak \ 
	label},\\
\sigma \ \ \ \mathrm{otherwise}.
\end{cases}
\end{equation}

$Pr(\hat{x_i},\hat{x_j})$ is a prior distribution in Bernoulli and $\lambda,\sigma\in [0,1]$. The prior distribution can be further learned by more complex distribution models like Gaussian mixture model \cite{rother2004grabcut}. We found this empirically predetermined Bernoulli distribution works well in our experiments.

\begin{figure}[t]
	\centering
	\includegraphics[width=0.8\linewidth]{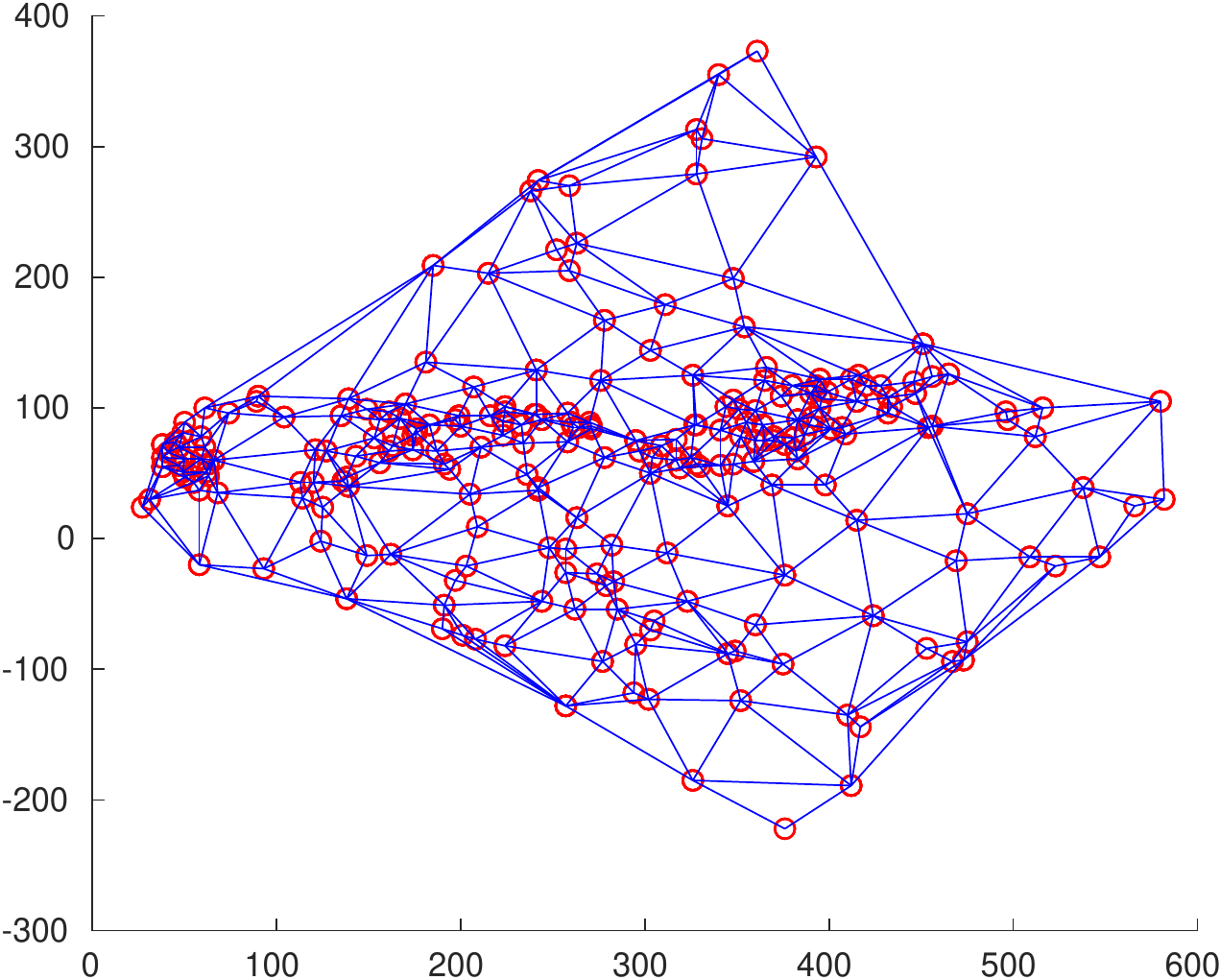}
	\caption{An example of constructing a neighbor system with Delaunay triangulation. Each point ${x_i}$ is in $\mathcal{{X}}$. In the case of high-dimensional data (e.g., feature-point pair is 4D in homography detection), the concatenated data is projected onto the first two principal axes extracted by PCA, and the neighbor system is constructed in this 2D plane. Distant edges are removed. }
	\label{fig:adjcent}
\end{figure}

\subsection{Proposal Sampling and Labeling with WA Data}
\label{sec:ps}
With the proposal sampling graph introduced, the \cz{update of proposals and the update of labeling results} can be realized by alternately sampling $d=\{d_{ij}\}$ and optimizing $\mathcal{L}$ under the random cluster framework \cite{pham2014random}, which solves $min_{d,\mathcal{L}}E$ instead of $min_{\mathcal{M},\mathcal{L}}E$. Notice that the sampling of $d$ and the optimization of $\mathcal{L}$ are respectively conducted with the two graphical models $\mathcal{\hat{G}}$ and $\mathcal{G}$ in our method. $\mathcal{G}$ is built with $\mathcal{X}$ for $\alpha$-expansion \cite{boykov1999fast}, as illustrated in Fig. \ref{fig:adjcent}. The two steps can be summarized as follows:\par
\vspace{2pt}\par
\noindent\textbf{\textit{Step (1)} $P(d \vert \mathcal{\hat{L}})$. Sampling $d$ with the current labeling result of WA data ($\mathcal{\hat{L}} \in \mathcal{L}$ on $\mathcal{\hat{G}}$):}\par
\begin{itemize}
	\item $P(d_{ij} = 1 \vert \hat{l_i}=\hat{l_j}):=w_{ij}$
	\item $P(d_{ij} = 1 \vert \hat{l_i}\neq \hat{l_j}):=0$
\end{itemize}

\noindent\textbf{\textit{Step (2)} $P( \mathcal{L} \vert d)$. Optimizing $\mathcal{L}$ by minimizing Eq. \ref{eq:EM} with the current $d$:}\par
\begin{itemize}
	\item A proposal is generated according to the sampled $d$ on $\mathcal{\hat{G}}$
	\item $\mathcal{L}$ is updated via $\alpha$-expansion by taking the new proposal into account.
\end{itemize}

The complexity term $O$ in Eq. \ref{eq:EM} is not involved in $\alpha$-expansion, as our algorithm does not follow the two-stage strategy \cite{Isack2012,lazic2009floss,yu2011global}: generating a huge number of random proposals and conducting labeling based on the proposals. In our method, one proposal is generated and probabilistically included or excluded under the framework of simulated annealing, which will not suffer from the complexity problem. The smooth term $S$ in Eq. \ref{eq:EM} follows the Potts model \cite{boykov1999fast} and is defined as $\sum_{(i,j)\in \mathcal{G}}{c_{ij}s_{ij}}$, and
\begin{equation}
s_{ij}=\begin{cases}
1 \ \ \ \textrm{if} \ l_i=l_j\\
0 \ \ \ \textrm{if} \ l_i\neq l_j
\end{cases}.   
\end{equation}

The smooth prior $c_{ij}$ can be defined with spatial prior, since closer points in the neighbor system can be more likely to fit the same model. For simplicity, we set the $c_{ij}$ as a fixed constant that only penalizes the discontinuity for each edge.

\textbf{\textit{Step (1)}} generates clusters of WA data, with each cluster indicating a model proposal. \textbf{\textit{Step (2)}} uses the proposals to perform labeling, and the labeling result in return encourages \textbf{\textit{Step (1)}} to connect the WA points which hold a same label. Obviously, this is a chicken-and-egg problem as the calculation in one of the two steps depends on the result of the other. Good labeling improves clustering and vice versa. An iterative algorithm is a realistic solution for getting rid of this situation. We modify the simulated annealing \cite{pham2014random} to further involve an optional subjective prior limitation by introducing a variable $nflag\in\{0,1\}$ to indicate whether the following assumption holds true ($nflag=1$) or not ($nflag=0$): the number of weak labels equals the number of embedded models (outlier model excluded). The whole procedure is listed in Alg. \ref{alg:WA-RCM}.

\begin{algorithm}[t]                      
	\caption{Geometric Multi-Model Fitting with Weak Annotations (G2MF-WA)}    
	\label{alg:WA-RCM}                          
	\begin{algorithmic}[1]         
		\REQUIRE {Data set ${\mathcal{X}}$, including WA data $\hat{\mathcal{X}}$. Proposal set $\Theta=\emptyset$, proposal pool $\mathcal{H}=\emptyset$, $\lambda=0.1$, $\sigma=0.1$, $h=10$, $c_{ij}=0.1$}, initial temperature $T$, $nflag\in\{0,1\}$, number of ground-truth labels $nlabels$ if $nflag=1$
		\ENSURE {Estimated models $\Theta$ and labels $\mathcal{L}$}
		\STATE{Construct $\mathcal{\hat{G}}$ and $\mathcal{G}$ with $\mathcal{\hat{X}}$ and $\mathcal{X}$ respectively}
		\IF{$nflag=0$} 
		\STATE{Sample $randn\in [0,1]$ uniformly}
		\IF{$randn<0.5$}
		\STATE{Update $w$ by Eq. \ref{eq:ourW} in $\mathcal{\hat{G}}$ and sample a $d$ (Sec. \ref{sec:psg})}
		\STATE{Update proposal set $\Theta$ by adding the new proposal $\theta_g$ generated from $d$, $\Theta'=\Theta \bigcup \theta_g$ (Sec. \ref{sec:ps})}
		\STATE{Add $\theta_g$ to $\mathcal{H}$}
		\ELSE
		\STATE{Update $\Theta$ by randomly removing one proposal $\theta_g$, $\Theta'=\Theta \backslash \theta_g$}
		\ENDIF
		\ELSIF{$nflag=1$}
		\IF{$\vert \Theta \vert < nlabels$}
		\STATE{Run step 5 $\sim$ step 7}
		\ELSE
		\STATE{Run step 9}
		\ENDIF
		\ENDIF
		\IF{$\vert\mathcal{H}\vert$>100}
		\STATE{$h=\lceil 0.1\times \vert \mathcal{H} \vert \rceil$}
		\ENDIF
		\STATE{Estimate $\mathcal{L}'$ with $\Theta'$ via $\alpha$-expansion on $\mathcal{G}$ (Sec. \ref{sec:ps})}
		\IF{$E(\mathcal{X},\Theta',\mathcal{L}')<E(\mathcal{X},\Theta,\mathcal{L})$}
		\STATE{$\Theta=\Theta', \mathcal{L}=\mathcal{L}'$}
		\ELSE
		\STATE{Run step 23 with probability $\exp{((E(\mathcal{X},\Theta,\mathcal{L})-E(\mathcal{X},\Theta',\mathcal{L}')}$)/T)}
		\STATE{$T:=0.99T$ and repeat from step 2 until $T\approx 0$}
		\ENDIF
		
	\end{algorithmic}
\end{algorithm}

\begin{figure}[t]
	\centering
	\subfigure[SA-RCM \cite{pham2014random}]{\includegraphics[width=0.48\linewidth]{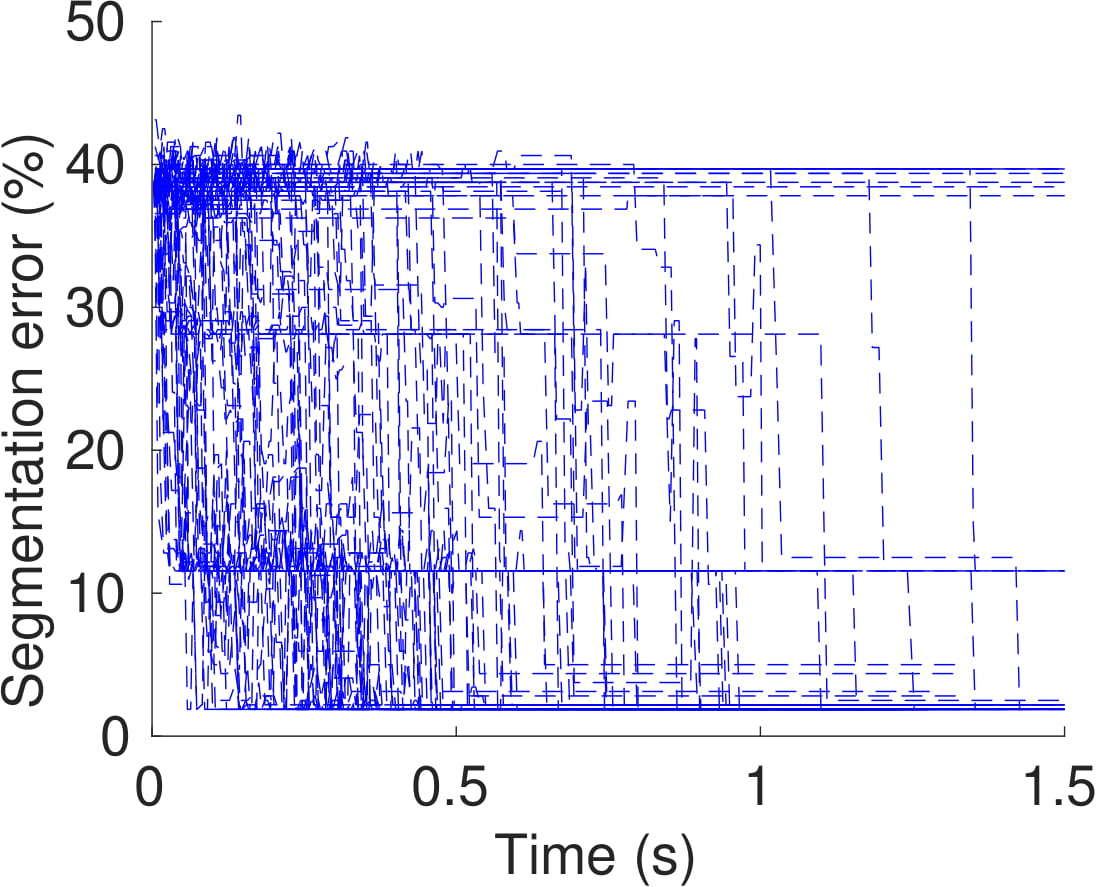}}
	\subfigure[G2MF-WA]{\includegraphics[width=0.48\linewidth]{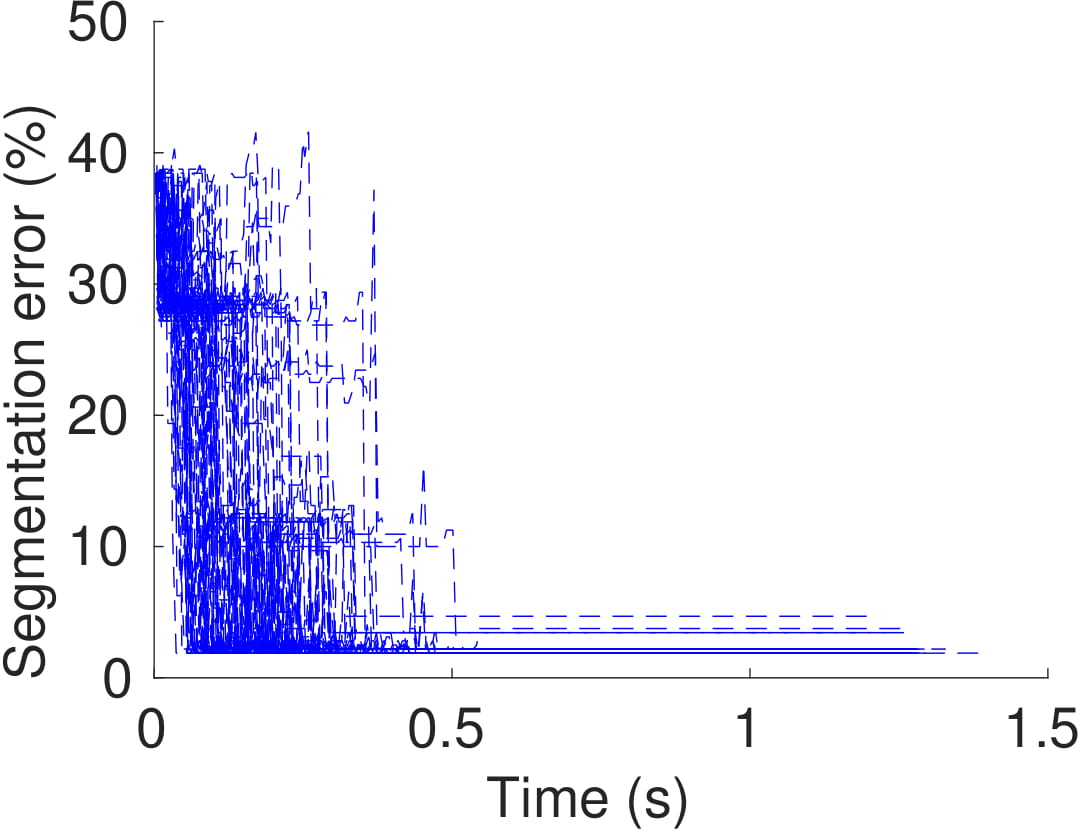}}
	
	\caption{Convergence analysis on the dataset \textit{hartley} from \cite{wong2011dynamic}. Dotted lines denote the convergence curves with respect to different random seeds. }
	\label{fig:convergence}
\end{figure}

To justify the improvement on convergence, we compare our method with SA-RCM \cite{pham2014random} in Fig. \ref{fig:convergence}. As the simulated annealing iterations are meta-heuristic, we run 100 times for each method with different random seeds. We observe from Fig. \ref{fig:convergence} that G2MF-WA \cz{(our method)} converges faster with lower segmentation errors in most of the trials than SA-RCM. It is easy to tell that G2MF-WA generally achieves convergence in less than 0.5 seconds with the aid of WA data, while SA-RCM still not converges within 1.5 seconds in some trials.


\section{Experimental Results}
We first explain the experimental setup, including the compared techniques and the involved parameters. Then we introduce three applications of our approach: multi-homography detection and two-view motion segmentation.

\subsection{Experimental setup}
We compared our method (denoted as G2MF-WA) with two state-of-the-art methods PEARL \cite{Isack2012} and SA-RCM \cite{pham2014random}. Note that the comparisons are not under the same problem setting, as we utilize additional weak annotated data which can be easily obtained. The purpose of the comparisons is to demonstrate that the weak annotations can help achieve better fitting results. Older methods which have been shown to be less accurate in \cite{pham2014random} are not included \cite{lazic2009floss,yu2011global,pham2011simultaneous}. Parameters of all the methods are carefully tuned based on the authors' implementations for best performance. The settings of each method are explained as follows.

\textbf{PEARL} \cite{delong2012fast}. As a typical two-stage method, PEARL generates a large proposal set ${\Theta}$ at once, followed by energy (Eq. \ref{eq:EM}) minimization procedure with complexity term. It is unlike SA-RCM and G2MF-WA which expand $\Theta$ sequentially from an empty set. The complexity term is formulated using label costs, and counts the number of unique labels in $\mathcal{L}$ and penalizes complex solutions. Minimization is realized by running $\alpha$-expansion iteratively, and the optimum solution after each iteration $\mathcal{L}'$ corresponds to an optimum $\Theta' \in \Theta$. $\Theta'$ is then refined with the labeling results $\mathcal{L}'$ and $\Theta$ is replaced with $\Theta'$. Iterations are repeated until convergence. It is obvious that the number and quality of proposals in the initial $\Theta$ significantly affect the final result. We set $\vert \Theta \vert=1000$ to ensure the accuracy. The minimum iteration number is set to 10 and the maximum iteration number is set to 20 as PEARL often converges within a few iterations.

\textbf{SA-RCM} \cite{pham2014random}. To control the convergence procedure in a practical way, the minimum iteration number is set to 500 and the maximum iteration number is set to 5000. The iteration is terminated when the energy changes in a small range over iterations. Unlike G2MF-WA, SA-RCM conducts proposal sampling and $\alpha$-expansion in the same adjacent graph over all data points, which induces more computational cost with Eq. \ref{eq:Auv} when the sizes of $\mathcal{X}$ and $\mathcal{G}$ grow.

\textbf{G2MF-WA (our method)}. G2MF-WA has a similar simulated annealing framework to SA-RCM. The hyperparameters are shown in Alg. \ref{alg:WA-RCM} and the number of iterations is set to be the same as SA-RCM. We generate simulated weak annotations instead of real manual annotations \cz{to facilitate the annotation controllability and enable large-scale comparisons.
}. The simulated weak annotations consist of two types of data: (1) $N_g$ points are randomly selected from each ground-truth label (except the outlier). (2) $N_o$ points from the ground-truth outlier are selected and assigned with other ground-truth labels randomly. For clarity, we design the following terms based on different weak annotation settings.
\begin{itemize}
	\item G2MF-WA-A: $N_g=5, N_o=5, \vert \mathcal{\hat{X}}\vert=10$
	\item G2MF-WA-B: $N_g=10, N_o=10, \vert \mathcal{\hat{X}}\vert=20$
	\item G2MF-WA-C: $N_g=10, N_o=5, \vert \mathcal{\hat{X}}\vert=15$
\end{itemize}

The segmentation error \cite{pham2014random} based on the ground-truth labeling is used as the evaluation criteria, which can be calculated for all the methods. All the experiments are conducted on an off-the-shelf PC with an Intel i7 CPU (3.6GHz) and 32GB RAM.

\subsection{Application 1: multi-homography detection}
Given two views of a scene, a number of feature points from these two images can be extracted and matched by feature matching techniques. Matched points can be further related by a $3 \times 3$ homography matrix if the points lie on the same planar structure. The goal of multi-homography detection is to recover such homography matrices from a set of matches. One homography corresponds to one model and incorrect matches correspond to the outlier model in our fitting algorithm. The DLT algorithm \cite{adbel1971direct} which requires at least four matches is employed for model generation, and the \cz{residual error} is calculated by Sampson distance. The full H part of the AdelaideRMF dataset \cite{wong2011dynamic} is used in this experiment for fairness purposes. The examples and statistical results are shown in Fig. \ref{fig:homoexample} and Tab. \ref{tab:1}, respectively. \cz{In Tab. \ref{tab:1}, colored cells represent top-3 cells in each row (same dataset, different methods), in terms of median error and average processing time. According to the total number of colored cells in each column, which is summarized in the last row, it is clear that the G2MF-WA-A converges the fastest and G2MF-WA-B and G2MF-WA-C achieve the lowest segmentation error}. The processing time includes both sampling and optimization time. In Fig. \ref{fig:homoexample}, according to the segmentation error in parentheses, we can find that either the optional assumption holds true (1st row) or not (2nd row), the WA data clearly contributes to improving the performance.

\begin{table*}[tb]
	\centering
	\caption{Median results over 100 trials on the multi-homography detection task with full AdelaideRMF dataset (H part). Darker colors represent lower errors (\%) and runtime (in seconds), denoted by fuchsia and cyan, respectively. Top-3 cells in each row are colored.}
	\begin{tabular}{|l|c|c|c|c|c|c|c|c|c|c|} \hline
		Method & \multicolumn{2}{|c|}{PEARL \cite{Isack2012}} & \multicolumn{2}{|c|}{SA-RCM \cite{pham2014random}} & \multicolumn{2}{|c|}{G2MF-WA-A} & \multicolumn{2}{|c|}{G2MF-WA-B} & \multicolumn{2}{|c|}{G2MF-WA-C}
		\\ \hline
		Dataset (\#labels)& Error & Time & Error & Time & Error & Time & Error & Time & Error & Time 
		\\ \hline \hline
		barrsmith (3) & \cellcolor{e2}{10.37} & 1.37 & \cellcolor{e2}{10.37} & \cellcolor{t3}{1.35} &  \cellcolor{e2}{10.37} &  \cellcolor{t2}{1.05} & \cellcolor{e1}{1.66} & 1.52 & \cellcolor{e1}{1.66} & \cellcolor{t1}{1.02} 
		\\ \hline
		bonhall (7) & \cellcolor{e1}{6.09} & \cellcolor{t1}{2.88} &  \cellcolor{e3}{8.80} & 4.11 &  \cellcolor{e2}{8.57} &  \cellcolor{t2}{3.75} & 9.08 & 3.95 & \cellcolor{e3}{8.80} & \cellcolor{t3}{3.93} 
		\\ \hline
		bonython (2)& \cellcolor{e1}{1.52} & 1.24 &  \cellcolor{e2}{2.27} & 1.05 &  \cellcolor{e1}{1.52} &  \cellcolor{t1}{0.81} & \cellcolor{e1}{1.52} & \cellcolor{t3}{1.01} & \cellcolor{e1}{1.52} & \cellcolor{t2}{0.84} 
		\\ \hline
		elderhalla (3)& 20.56 & 1.40 &  \cellcolor{e3}{6.54} & 1.02 & 7.01 &  \cellcolor{t1}{0.92} & \cellcolor{e1}{5.61} & \cellcolor{t3}{0.97} & \cellcolor{e2}{6.07} & \cellcolor{t2}{0.95} 
		\\ \hline
		elderhallb (4)& 21.57 & 1.43 &  6.27 &  \cellcolor{t1}{0.95} &  \cellcolor{e3}{5.88} &  \cellcolor{t2}{1.01} & \cellcolor{e1}{5.10} & 1.11 & \cellcolor{e2}{5.29} & \cellcolor{t3}{1.08} 
		\\ \hline
		hartley (3)& \cellcolor{e2}{2.19} & 1.66 &  \cellcolor{e1}{1.88} &  1.28 &  \cellcolor{e2}{2.19} &  \cellcolor{t1}{1.08} & \cellcolor{e2}{2.19} & \cellcolor{t3}{1.16} & \cellcolor{e2}{2.19} & \cellcolor{t2}{1.15} 
		\\ \hline
		johnsona (5)& \cellcolor{e3}{7.51} & 1.61 &  \cellcolor{e2}{3.49} &  \cellcolor{t1}{1.27} & \cellcolor{e3}{7.51} &  \cellcolor{t2}{1.31} & \cellcolor{e1}{3.22} & 1.46 & \cellcolor{e1}{3.22} & \cellcolor{t3}{1.42} 
		\\ \hline
		johnsonb (8)& 14.10 & \cellcolor{t1}{2.13} &  \cellcolor{e2}{9.86} &  \cellcolor{t2}{2.16} & 14.33 &  \cellcolor{t3}{2.37} & \cellcolor{e1}{9.40} & 2.54 & \cellcolor{e3}{10.02} & 2.57 
		\\ \hline
		ladysymon (3)& \cellcolor{e1}{4.64} & 1.33 &  \cellcolor{e2}{5.06} &  \cellcolor{t1}{0.93} &  \cellcolor{e1}{4.64} &  \cellcolor{t2}{0.96} & \cellcolor{e1}{4.64} & \cellcolor{t3}{0.99} & \cellcolor{e1}{4.64} & 1.01 
		\\ \hline
		library (3)& \cellcolor{e2}{3.26} & 1.30 &  \cellcolor{e2}{3.26} &  \cellcolor{t1}{0.88} &  \cellcolor{e1}{2.79} &  \cellcolor{t2}{0.90} & \cellcolor{e1}{2.79} & \cellcolor{t3}{0.94} & \cellcolor{e1}{2.79} & 0.96 
		\\ \hline
		mc1 (6)& 10.90 & \cellcolor{t1}{4.73} &  \cellcolor{e1}{4.30} &  9.11 &  5.36 &  \cellcolor{t2}{7.71} & \cellcolor{e2}{4.51} & \cellcolor{t3}{8.13} & \cellcolor{e3}{4.87} & 8.22 
		\\ \hline
		mc3 (7)& 29.92 & \cellcolor{t1}{4.80} &  4.69 & 9.22 &  \cellcolor{e3}{3.88} &  \cellcolor{t2}{8.38} & \cellcolor{e2}{2.93} & \cellcolor{t3}{8.42} & \cellcolor{e1}{2.62} & 8.51 
		\\ \hline
		napiera (3)& \cellcolor{e3}{17.22} & 1.58 & 18.05 &  1.40 & \cellcolor{e1}{12.58} &  \cellcolor{t1}{1.09} & \cellcolor{e2}{13.25} & \cellcolor{t2}{1.12} & \cellcolor{e2}{13.25} & \cellcolor{t3}{1.20} 
		\\ \hline
		napierb (4)& \cellcolor{e2}{17.37} & 1.44 & 18.92 &  \cellcolor{t1}{1.00} & 20.08 &  \cellcolor{t2}{1.07} & \cellcolor{e1}{15.44} & \cellcolor{t3}{1.12} & \cellcolor{e3}{17.95} & 1.15 
		\\ \hline
		neem (4)& 7.47 & 1.41 &  \cellcolor{e3}{5.39} &  \cellcolor{t1}{0.89} &  6.43 &  \cellcolor{t2}{0.99} & \cellcolor{e2}{4.98} & \cellcolor{t3}{1.04} & \cellcolor{e1}{4.56} & 1.06 
		\\ \hline
		nese (3)& \cellcolor{e1}{0.79} & 1.38 &  \cellcolor{e1}{0.79} &  \cellcolor{t1}{0.86} &  \cellcolor{e1}{0.79} &  \cellcolor{t2}{0.92} & \cellcolor{e1}{0.79} & 1.45 & \cellcolor{e1}{0.79} & \cellcolor{t3}{1.01} 
		\\ \hline
		oldclassicswing (3)& \cellcolor{e1}{1.06} & 1.65 &  \cellcolor{e1}{1.06} &  \cellcolor{t1}{1.16} &  \cellcolor{e1}{1.06} &  \cellcolor{t1}{1.16} & \cellcolor{e1}{1.06} & \cellcolor{t2}{1.23} & \cellcolor{e1}{1.06} & \cellcolor{t3}{1.29} 
		\\ \hline
		physics (2)& \cellcolor{e1}{19.81} & 1.03 & \cellcolor{e3}{26.42} &  0.88 & \cellcolor{e1}{19.81} &  \cellcolor{t1}{0.72} & \cellcolor{e2}{23.58} & \cellcolor{t3}{0.77} & 27.36 & \cellcolor{t2}{0.75} 
		\\ \hline
		raglan (12)& \cellcolor{e3}{17.73} & \cellcolor{t1}{6.07} & 45.48 & 15.17 & 42.46 & \cellcolor{t2}{14.79} & \cellcolor{e1}{10.23} & \cellcolor{t3}{15.03} & \cellcolor{e2}{10.90} & 15.08 
		\\ \hline
		sene (3)& \cellcolor{e1}{1.20} & 1.39 &  \cellcolor{e1}{1.20} &  \cellcolor{t1}{0.91} &  \cellcolor{e1}{1.20} &  \cellcolor{t1}{0.91} & \cellcolor{e1}{1.20} & \cellcolor{t3}{1.13} & \cellcolor{e2}{1.60} & \cellcolor{t2}{0.99} 
		\\ \hline
		unihouse (6)& 32.17 & \cellcolor{t1}{5.45} &  \cellcolor{e3}{4.32} & 10.18 &  4.70 &  \cellcolor{t2}{8.37} & \cellcolor{e2}{2.50} & 8.80 & \cellcolor{e1}{2.40} & \cellcolor{t3}{8.73} 
		\\ \hline
		unionhouse (6)& 38.15 & \cellcolor{t1}{5.25} &  \cellcolor{e3}{4.17} & 10.19 &  \cellcolor{e2}{2.74} &  \cellcolor{t2}{8.59} & \cellcolor{e1}{2.54} & \cellcolor{t3}{8.72} & \cellcolor{e1}{2.54} & \cellcolor{t3}{8.72} 
		\\ \hline \hline
		\cz{\#colored cells in each column}& 14 & 7 & 17 & 11 &  15 &  \textbf{22} & \textbf{21} & 15 & \textbf{21} & 14 
		\\ \hline
	\end{tabular}
	\label{tab:1}
\end{table*}

\begin{table*}[tb]
	\centering
	\caption{Median results over 100 trials on the two-view motion segmentation task with full AdelaideRMF dataset (F part). Darker colors represent lower errors (\%) and runtime (in seconds), denoted by fuchsia and cyan, respectively. Top-3 cells in each row are colored.}
	\begin{tabular}{|l|c|c|c|c|c|c|c|c|c|c|} \hline
		Method & \multicolumn{2}{|c|}{PEARL \cite{Isack2012}} & \multicolumn{2}{|c|}{SA-RCM \cite{pham2014random}} & \multicolumn{2}{|c|}{G2MF-WA-A} & \multicolumn{2}{|c|}{G2MF-WA-B} & \multicolumn{2}{|c|}{G2MF-WA-C}
		\\ \hline
		Dataset (\#labels)& Error & Time & Error & Time & Error & Time & Error & Time & Error & Time 
		\\ \hline \hline
		biscuit (2)&  \cellcolor{e2}{0.30} &  \cellcolor{t1}{1.01} &  \cellcolor{e1}{0.00} & 1.36 & - & - &  \cellcolor{e3}{0.61} &  \cellcolor{t3}{1.07} &  \cellcolor{e3}{0.61} &  \cellcolor{t2}{1.05} 
		\\ \hline
		biscuitbook (3)&  \cellcolor{e2}{1.47} & \cellcolor{t1}{1.04} &  2.93 & \cellcolor{t1}{1.04} & - & - &  \cellcolor{e1}{1.17} &  \cellcolor{t3}{1.17} &  \cellcolor{e3}{2.35} &  \cellcolor{t2}{1.12} 
		\\ \hline
		biscuitbookbox (4)& \cellcolor{e1}{1.93} &  \cellcolor{t2}{0.98} &  \cellcolor{e2}{2.70} &  \cellcolor{t1}{0.89} & - & - &  \cellcolor{e1}{1.93} &  1.07 &  \cellcolor{e1}{1.93} &  \cellcolor{t3}{1.01} 
		\\ \hline
		boardgame (4)& \cellcolor{e2}{16.49} &  \cellcolor{t3}{1.00} & \cellcolor{e3}{16.85} &  \cellcolor{t1}{0.95} & - & - & \cellcolor{e1}{16.13} & 1.15 & 17.20 &  \cellcolor{t2}{1.09} 
		\\ \hline
		book (2)&  \cellcolor{e2}{1.07} &  \cellcolor{t3}{0.87} &  \cellcolor{e1}0.53 &  0.97 & - & - &  \cellcolor{e1}{0.53} &  \cellcolor{t2}{0.83} &  \cellcolor{e1}{0.53} &  \cellcolor{t1}{0.78} 
		\\ \hline
		breadcartoychips (5)& 29.11 &  \cellcolor{t2}{0.95} &  \cellcolor{e1}{6.96} &  \cellcolor{t1}{0.92} & - & - & \cellcolor{e3}{13.71} &  1.04 &  \cellcolor{e2}{8.86} &  \cellcolor{t3}{1.00} 
		\\ \hline
		breadcube (3)&  \cellcolor{e2}{5.58} &  \cellcolor{t3}{0.96} & 23.55 &  \cellcolor{t1}{0.83} & - & - &  \cellcolor{e3}{7.64} &  \cellcolor{t2}{0.85} &  \cellcolor{e1}{3.72} &  \cellcolor{t1}{0.83} 
		\\ \hline
		breadcubechips (4)& \cellcolor{e3}{14.78} &  \cellcolor{t2}{0.93} &  \cellcolor{e1}{7.39} &  \cellcolor{t1}{0.85} & - & - &  \cellcolor{e2}{9.57} &  0.99 &  \cellcolor{e1}{7.39} &  \cellcolor{t3}{0.94} 
		\\ \hline
		breadtoy (3)&  \cellcolor{e3}{3.82} & \cellcolor{t3}{1.01} &  \cellcolor{e2}{3.65} &  \cellcolor{t1}{0.97} & - & - &  \cellcolor{e3}{3.82} &  1.05 &  \cellcolor{e1}{3.30} &  \cellcolor{t2}{1.00} 
		\\ \hline
		breadtoycar (4)& 31.33 &  \cellcolor{t2}{0.86} &  \cellcolor{e1}{9.04} &  \cellcolor{t1}{0.76} & - & - & \cellcolor{e3}{18.37} &  0.92 &  \cellcolor{e2}{9.34} &  \cellcolor{t3}{0.87} 
		\\ \hline
		carchipscube (4)& 15.15 &  \cellcolor{t3}{0.87} & \cellcolor{e3}{12.12} &  \cellcolor{t1}{0.73} & - & - &  \cellcolor{e2}{3.64} &  0.88 &  \cellcolor{e1}{2.42} &  \cellcolor{t2}{0.85} 
		\\ \hline
		cube (2)&  \cellcolor{e2}{1.66} &  \cellcolor{t2}{1.00} & 10.93 &  \cellcolor{t3}{1.18} & - & - &  \cellcolor{e3}{2.65} &  \cellcolor{t2}{1.00} &  \cellcolor{e1}{1.32} &  \cellcolor{t1}{0.97} 
		\\ \hline
		cubebreadtoychips (5)& 21.41 &  \cellcolor{t1}{1.08} &  \cellcolor{e2}{8.10} &  \cellcolor{t2}{1.13} & - & - &  \cellcolor{e1}{7.80} &  1.25 &  \cellcolor{e3}{9.02} &  \cellcolor{t3}{1.20} 
		\\ \hline
		cubechips (3)& 7.75 &  \cellcolor{t2}{0.99} &  \cellcolor{e1}{3.52} &  \cellcolor{t1}{0.94} & - & - &  \cellcolor{e3}{5.63} &  \cellcolor{t3}{1.05} &  \cellcolor{e2}{4.05} &  \cellcolor{t2}{0.99} 
		\\ \hline
		cubetoy (3)&  7.23 &  \cellcolor{t3}{0.95} &  \cellcolor{e3}{5.62} &  \cellcolor{t1}{0.82} & - & - &  \cellcolor{e2}{5.22} &  0.97 &  \cellcolor{e1}{4.02} &  \cellcolor{t2}{0.90} 
		\\ \hline
		dinobooks (4)& 20.83 &  \cellcolor{t1}{1.09} & \cellcolor{e3}{17.08} &  \cellcolor{t3}{1.24} & - & - & \cellcolor{e2}{16.25} &  1.25 & \cellcolor{e1}{12.50} &  \cellcolor{t2}{1.20} 
		\\ \hline
		game (2)&  \cellcolor{e2}{1.72} &  \cellcolor{t3}{0.90} &  4.72 &  1.05 & - & - &  \cellcolor{e3}{3.43} &  \cellcolor{t2}{0.87} &  \cellcolor{e1}{0.86} &  \cellcolor{t1}{0.83} 
		\\ \hline
		gamebiscuit (3)&  \cellcolor{e3}{1.52} &  \cellcolor{t2}{1.04} &  6.10 &  \cellcolor{t1}{1.01} & - & - &  \cellcolor{e2}{1.22} &  1.16 &  \cellcolor{e1}{0.61} &  \cellcolor{t3}{1.05} 
		\\ \hline
		toycubecar (4)& \cellcolor{e3}{23.50} &  \cellcolor{t2}{0.89} & \cellcolor{e1}{13.00} &  \cellcolor{t1}{0.73} & - & - & \cellcolor{e1}{13.00} &  \cellcolor{t3}{0.92} & \cellcolor{e2}{14.00} &  \cellcolor{t2}{0.89} 
		\\ \hline \hline
		\cz{\#colored cells in each column}& 12 & \textbf{19} & 14 & 16 & - & - & \textbf{19} & 8 & 18 & \textbf{19} 
		\\ \hline
	\end{tabular}
	\label{tab:2}
\end{table*}

\subsection{Application 2: two-view motion segmentation}
\begin{figure*}[tb]
	\centering
	\subfigure[Ground-truth]{\includegraphics[width=0.18\linewidth]{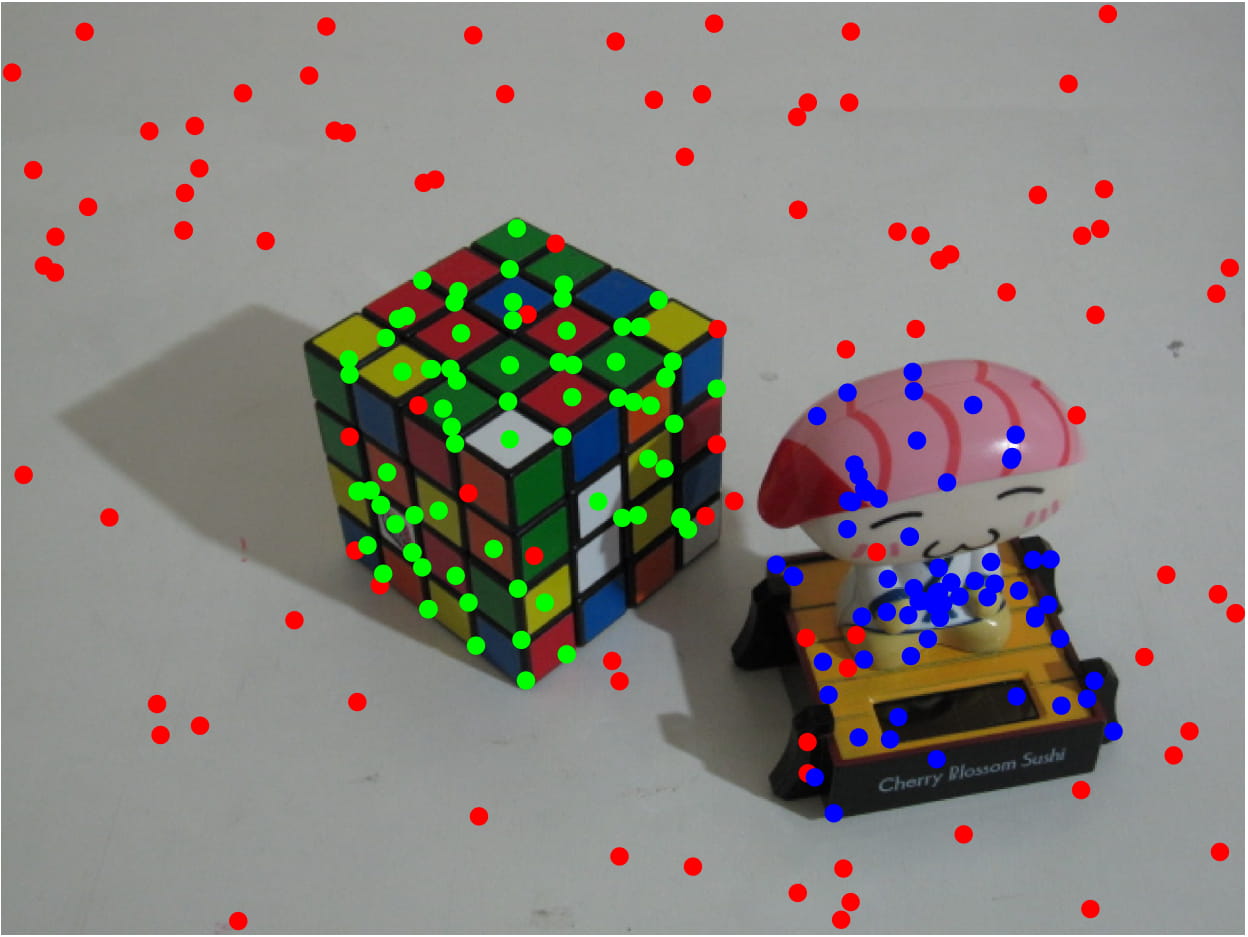}}
	\subfigure[Result of PEARL (7.2\%)]{\includegraphics[width=0.18\linewidth]{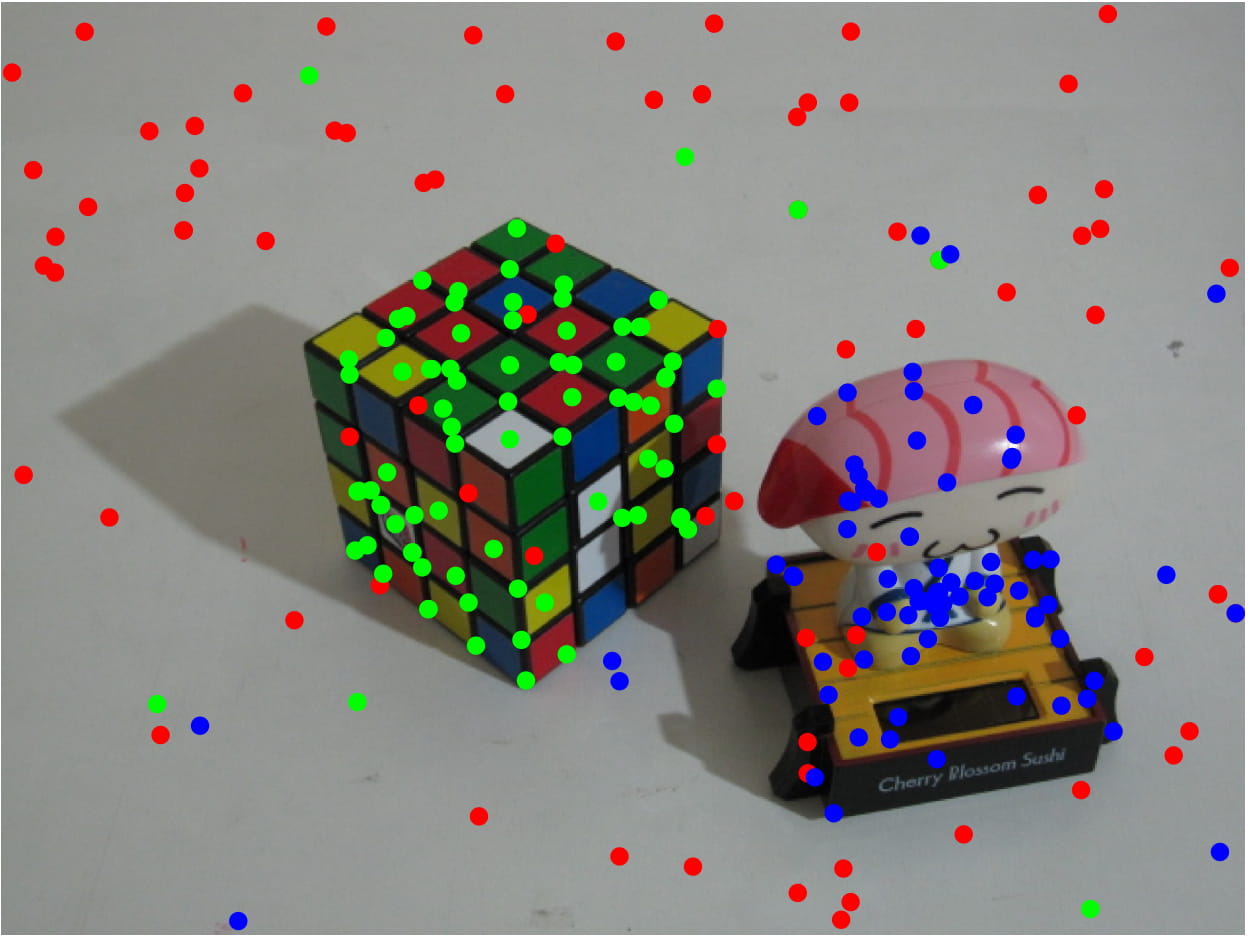}}
	\subfigure[Result of SA-RCM (5.2\%)]{\includegraphics[width=0.18\linewidth]{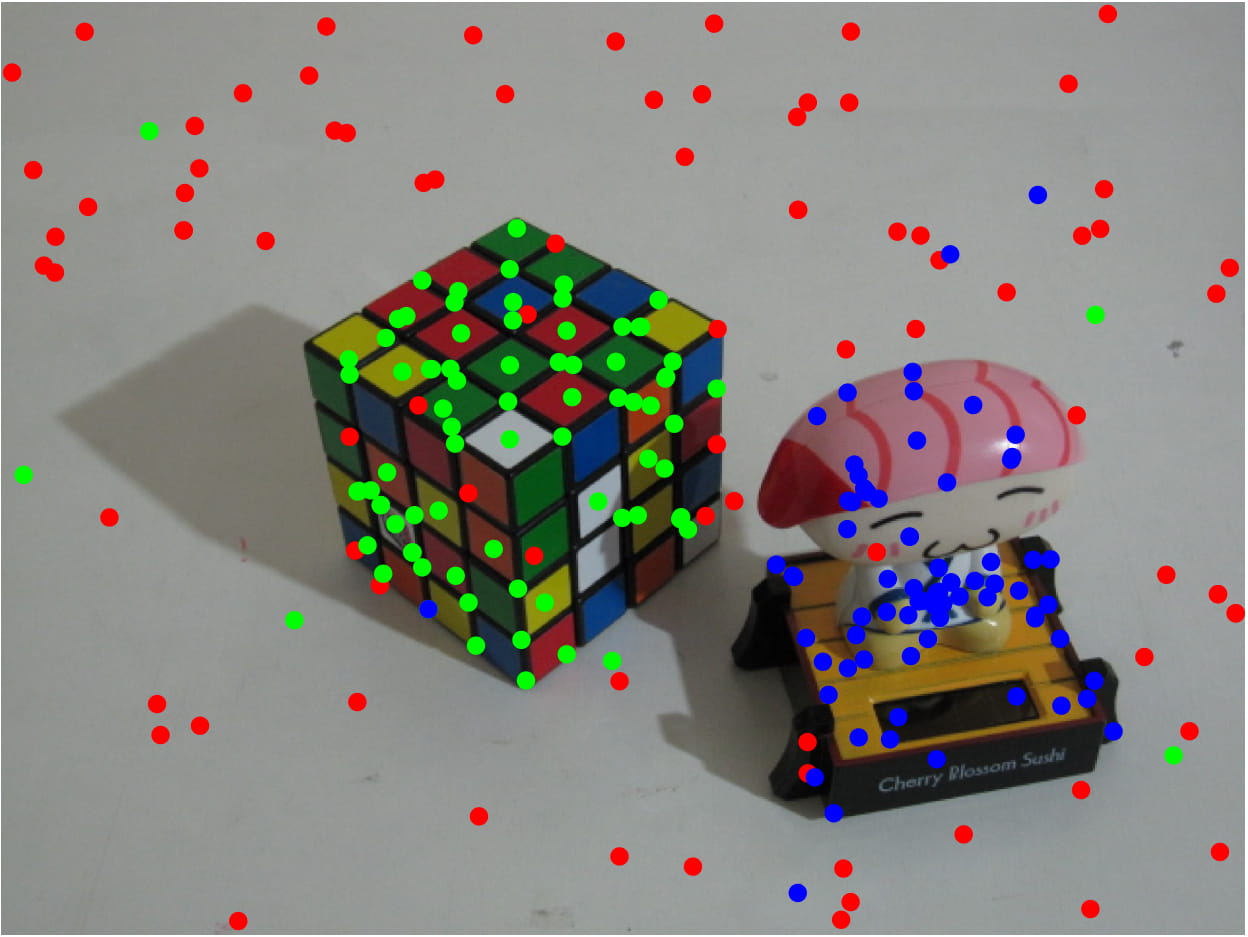}}
	\subfigure[Weak annotations]{\includegraphics[width=0.18\linewidth]{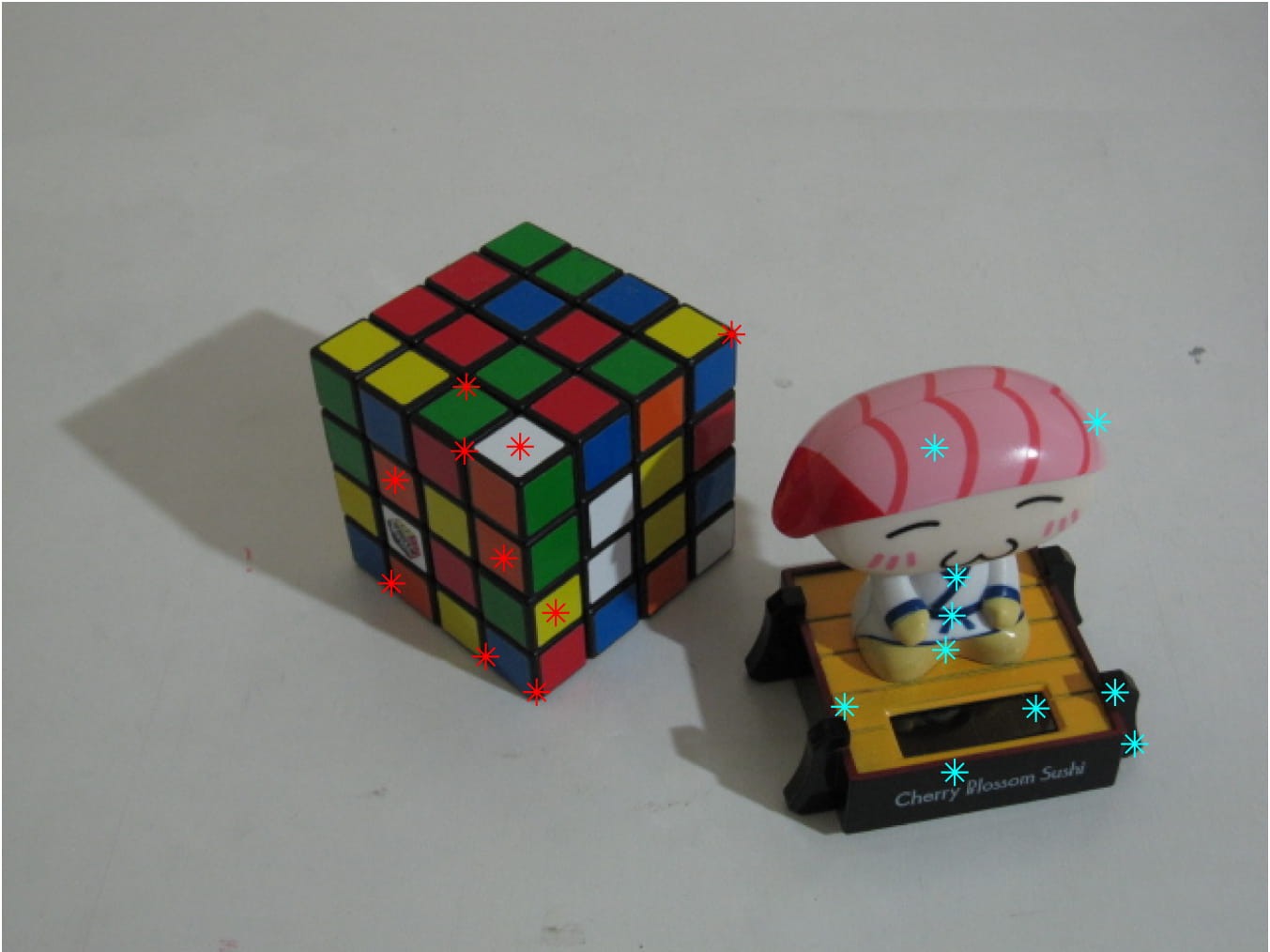}}
	\subfigure[Result of G2MF-WA (3.2\%)]{\includegraphics[width=0.18\linewidth]{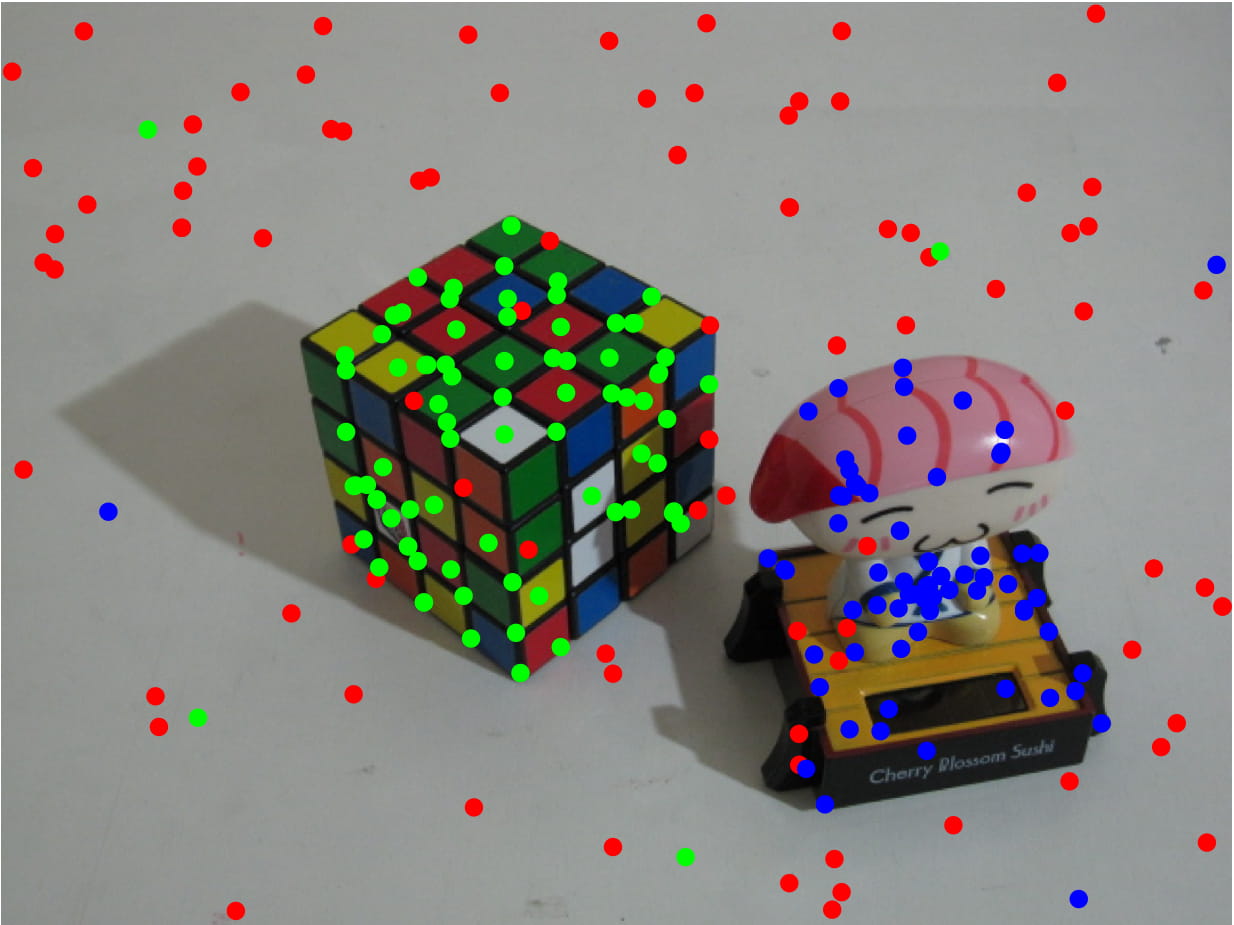}}\\
	\subfigure[Ground-truth]{\includegraphics[width=0.18\linewidth]{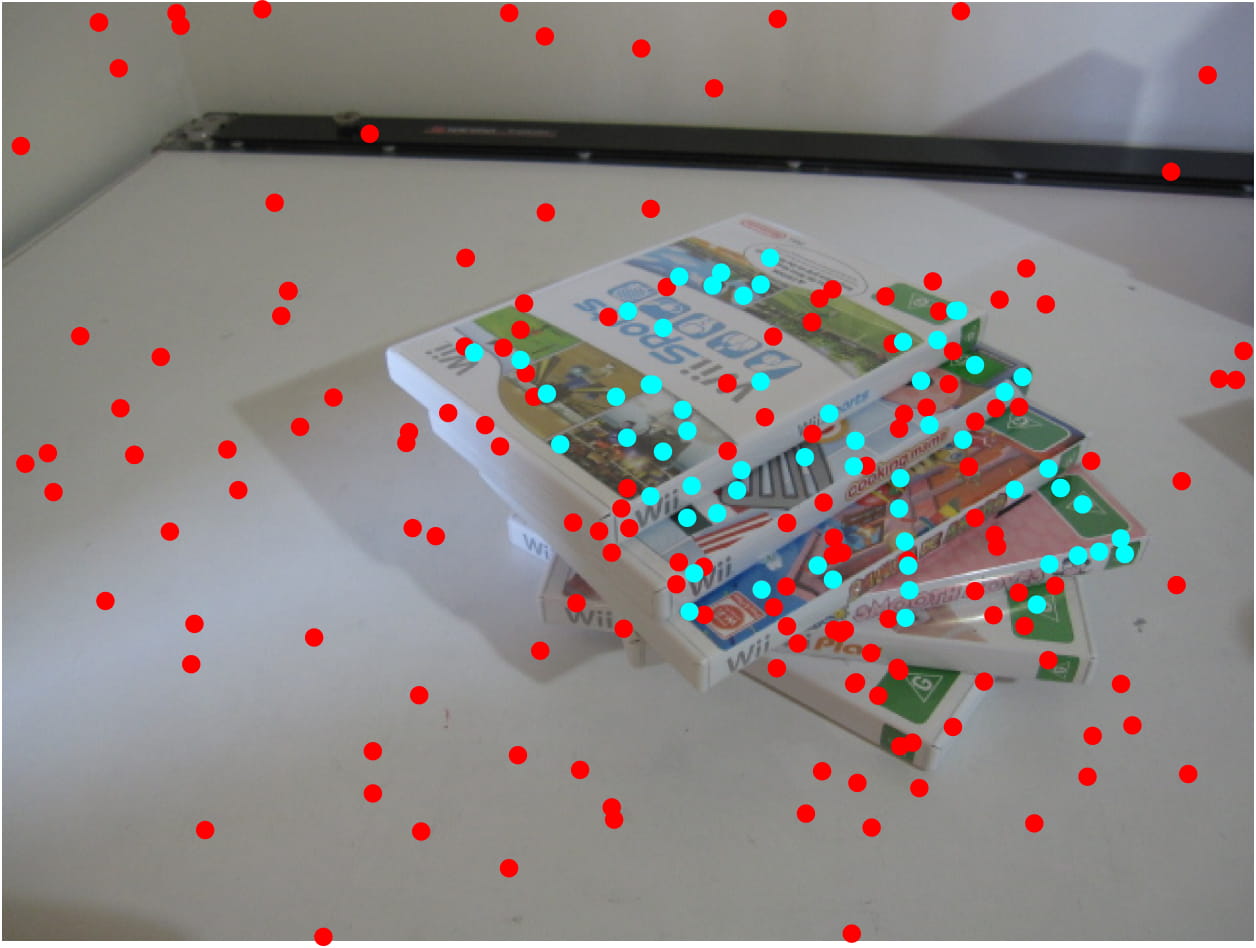}}
	\subfigure[Result of PEARL (5.2\%)]{\includegraphics[width=0.18\linewidth]{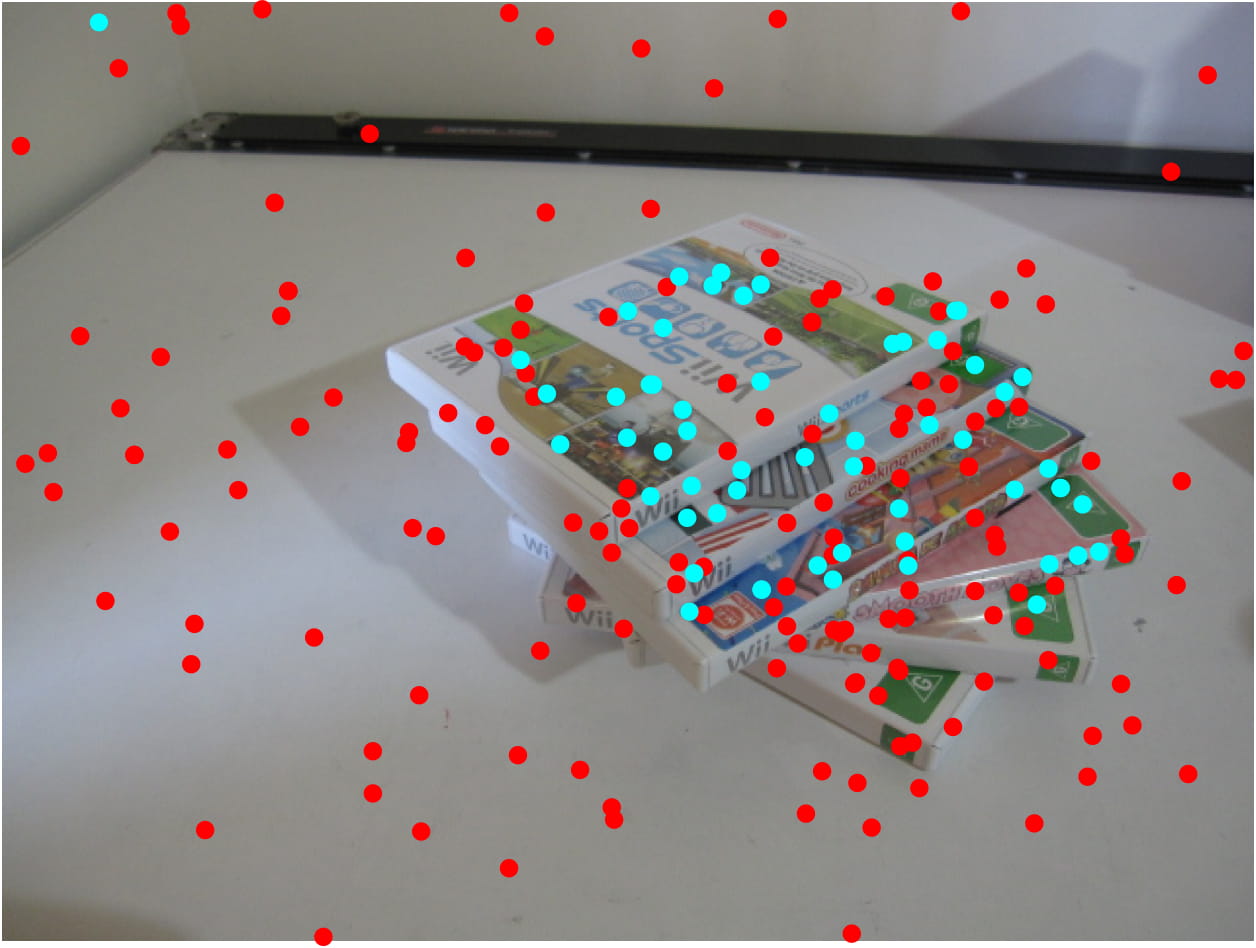}}
	\subfigure[Result of SA-RCM (4.3\%)]{\includegraphics[width=0.18\linewidth]{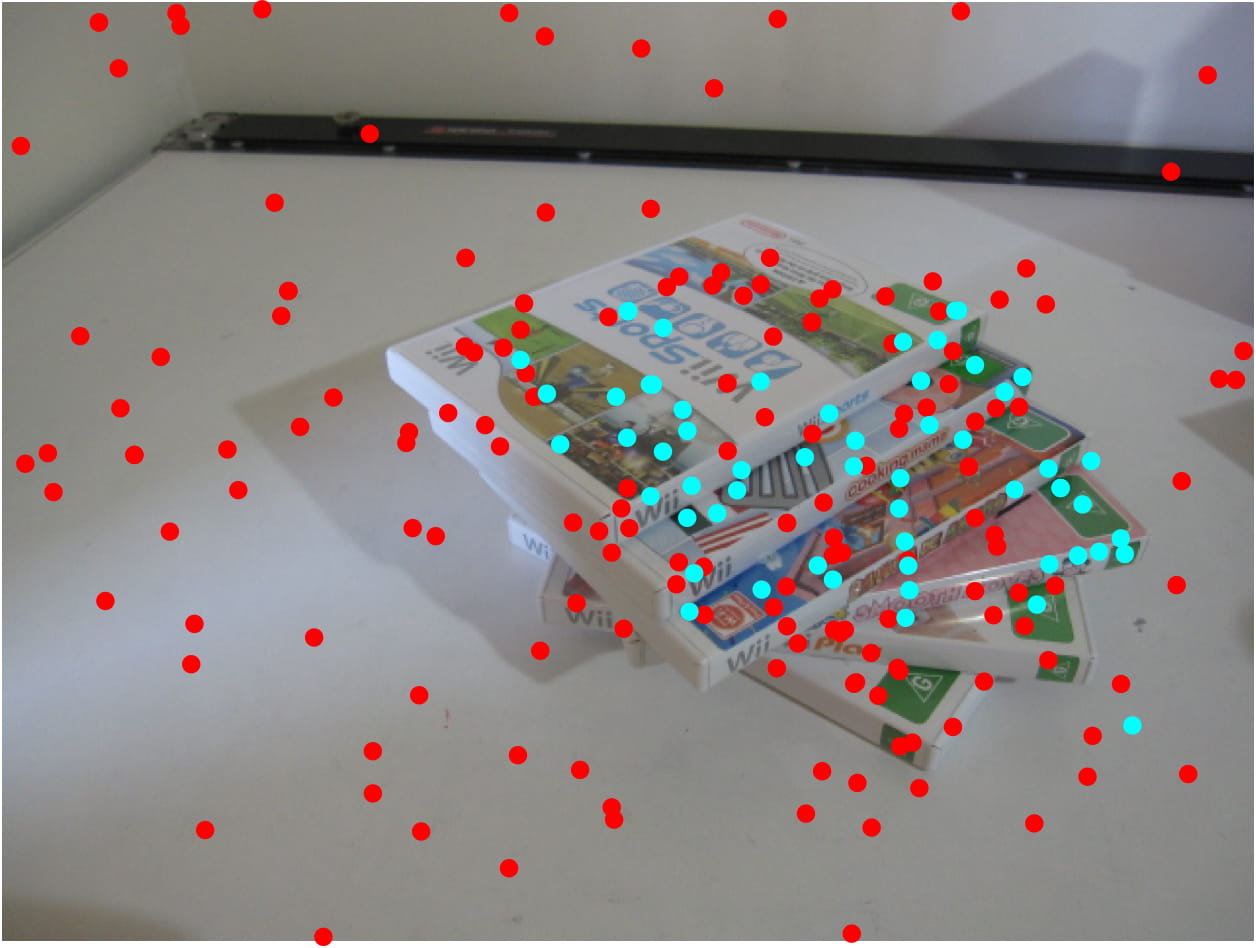}}
	\subfigure[Weak annotation]{\includegraphics[width=0.18\linewidth]{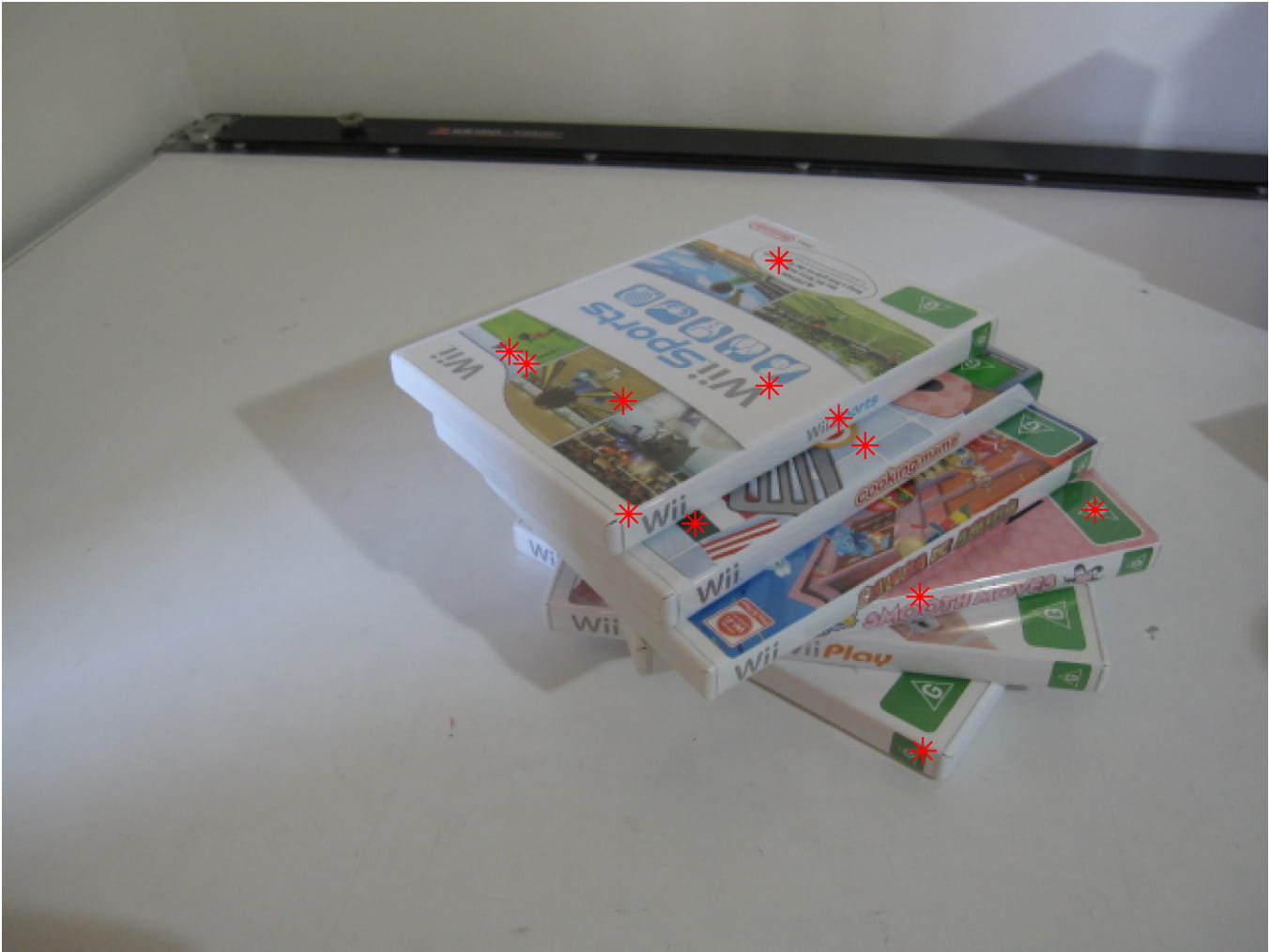}}
	\subfigure[Result of G2MF-WA (0.9\%)]{\includegraphics[width=0.18\linewidth]{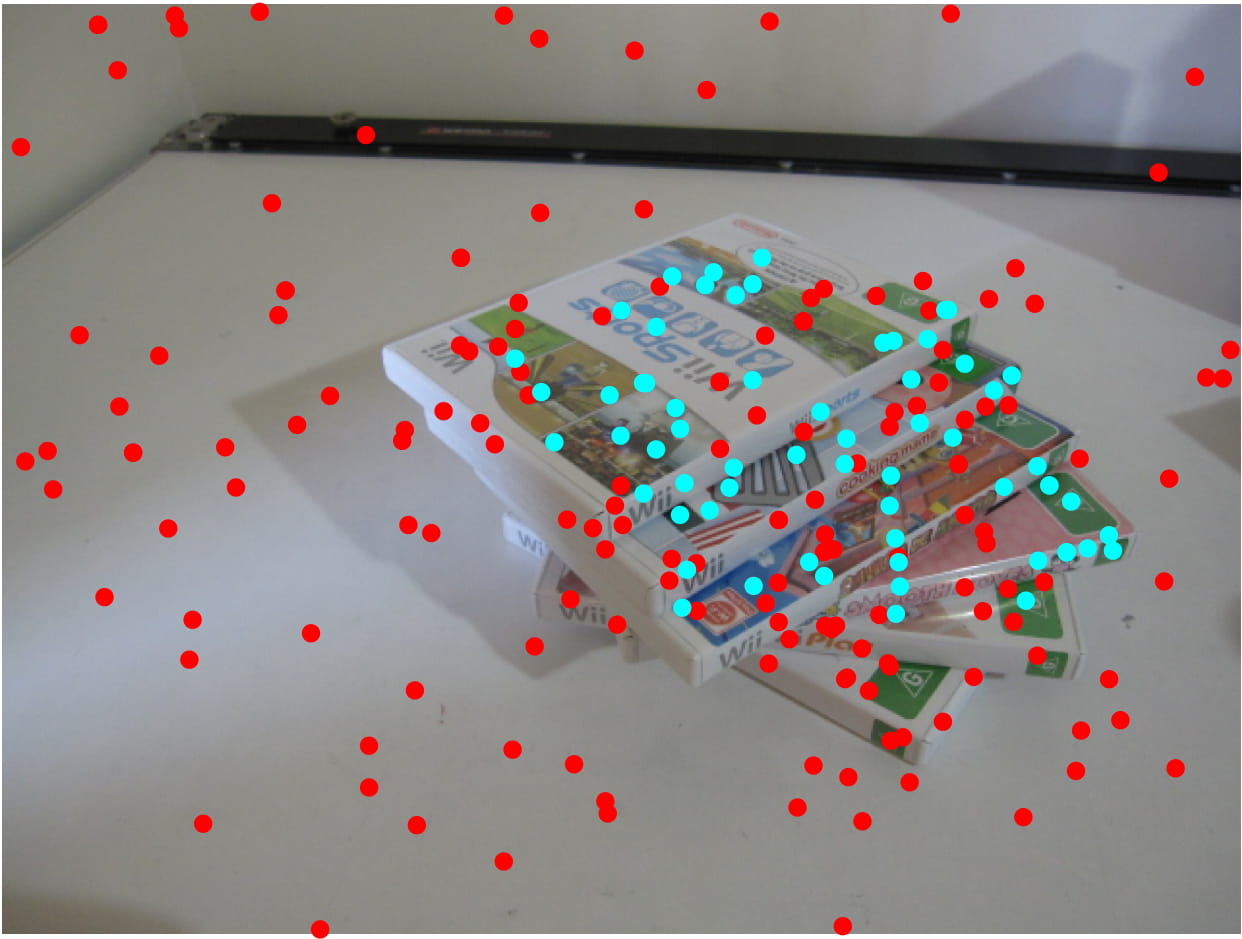}}
	
	\caption{Examples of the two-view motion segmentation. The first row (\textit{cubetoy}) shows the situation when the number of weak labels in (d) equals the number of ground-truth labels in (a) \cz{without considering the outlier label (red circle in (a)), i.e., 2 labels in (d) and 3 labels in (a) including the outlier.}. The second row (\textit{game}) shows the situation when only one label exists. In (d) and (i), \cz{WA data points are shown}. }
	\label{fig:fundexample}
\end{figure*}

Given two views of a scene and feature point matches, the goal of two-view motion segmentation is to estimate motion models modeled by $3\times3$ fundamental matrices and simultaneously the labeling. Points in a match are supposed to perform the same motion (usually on the same object or background). Outliers correspond to incorrect matches. The full F part of the AdelaideRMF dataset \cite{wong2011dynamic} is employed for fairness. The examples and statistical results are shown in Fig. \ref{fig:fundexample} and Tab. \ref{tab:2}, respectively. The DLT algorithm which requires at least eight matches is adopted for model generation, and the \cz{residual error} is calculated by Sampson distance. In the case of G2MF-WA-A in Tab. \ref{tab:2}, the results are unavailable as it is impossible to sample correct proposals when $N_g=5$, which is smaller than $8$. \cz{We can clearly observe from Tab. \ref{tab:2} that G2MF-WA-B achieves the lowest error yet performs the least efficiently, while G2MF-WA-C performs fastest and is also competitive in achieving low error}.

\begin{figure*}[tb]
	\centering
	\subfigure[Ground-truth]{\includegraphics[width=0.18\linewidth]{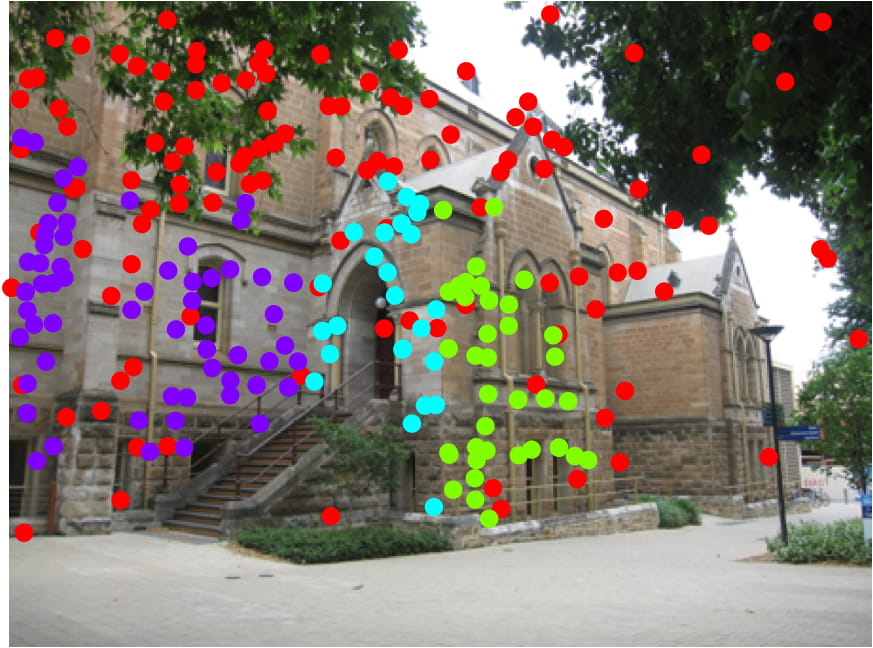}}
	\subfigure[Result of PEARL (22.4\%)]{\includegraphics[width=0.18\linewidth]{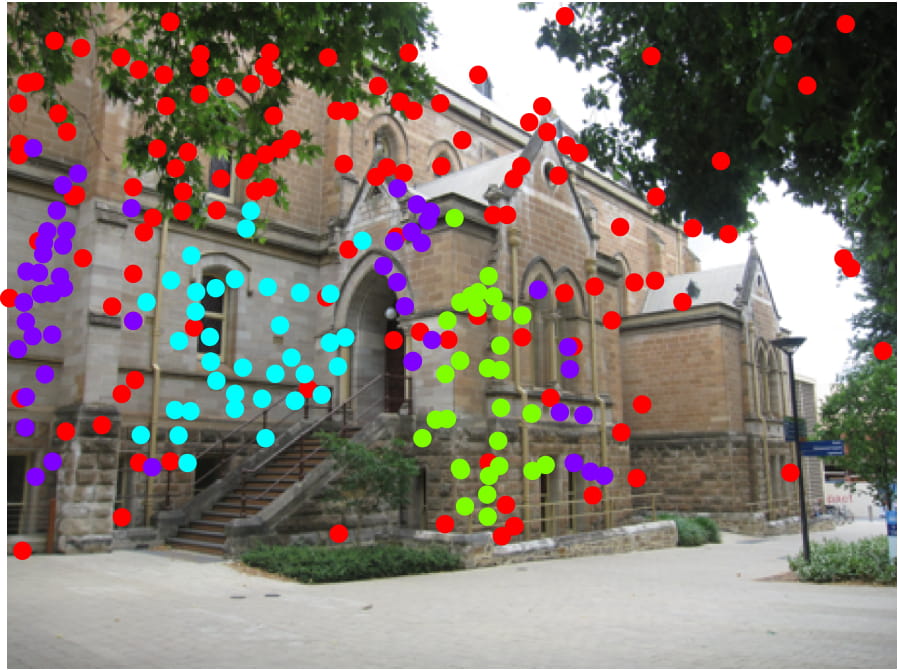}}
	\subfigure[Result of SA-RCM (6.7\%)]{\includegraphics[width=0.18\linewidth]{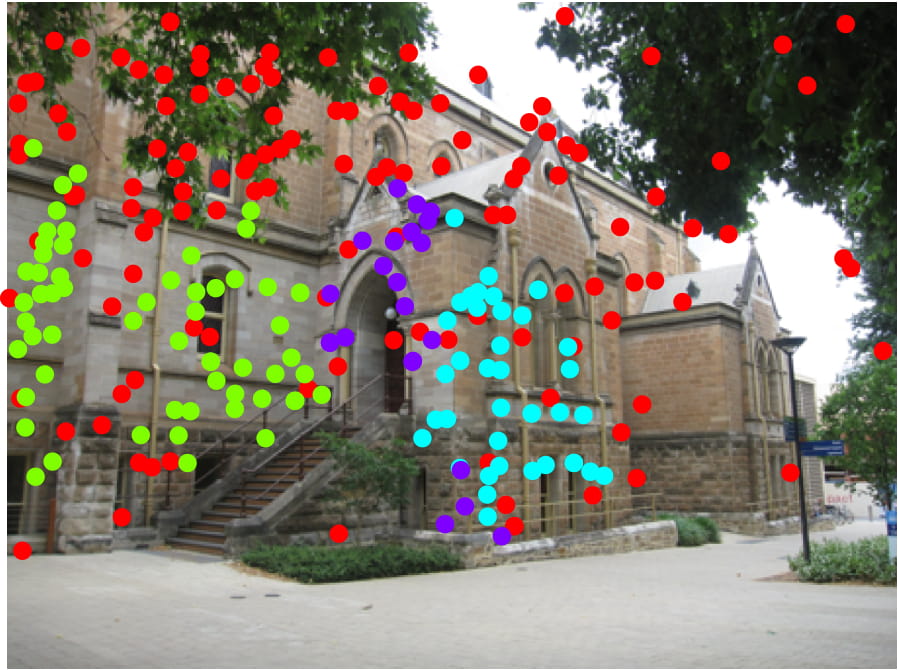}}
	\subfigure[Weak annotations]{\includegraphics[width=0.18\linewidth]{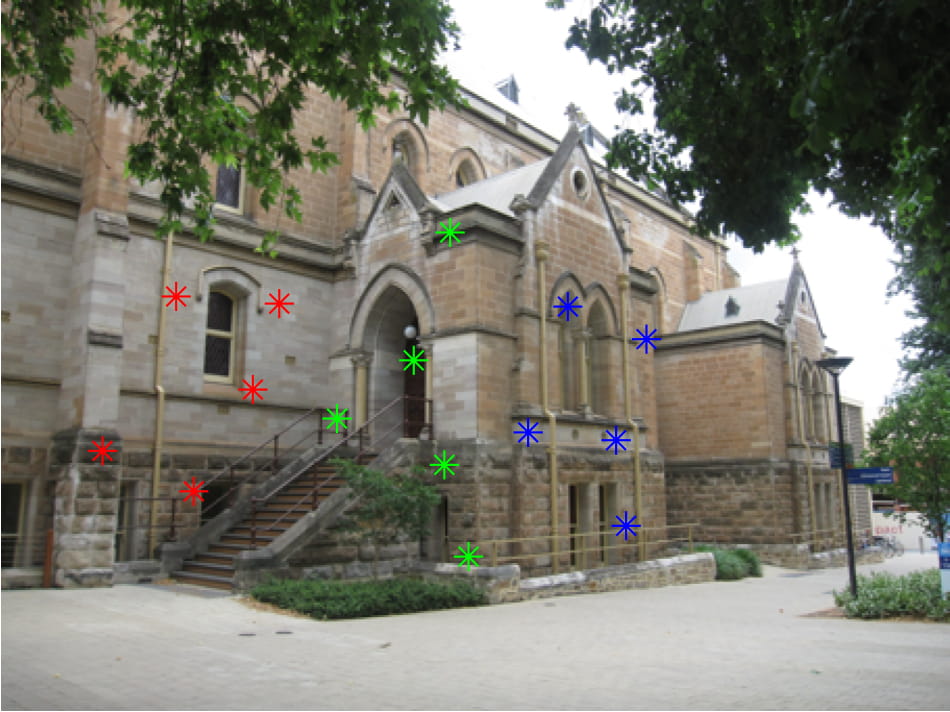}}
	\subfigure[Result of G2MF-WA (5.1\%)]{\includegraphics[width=0.18\linewidth]{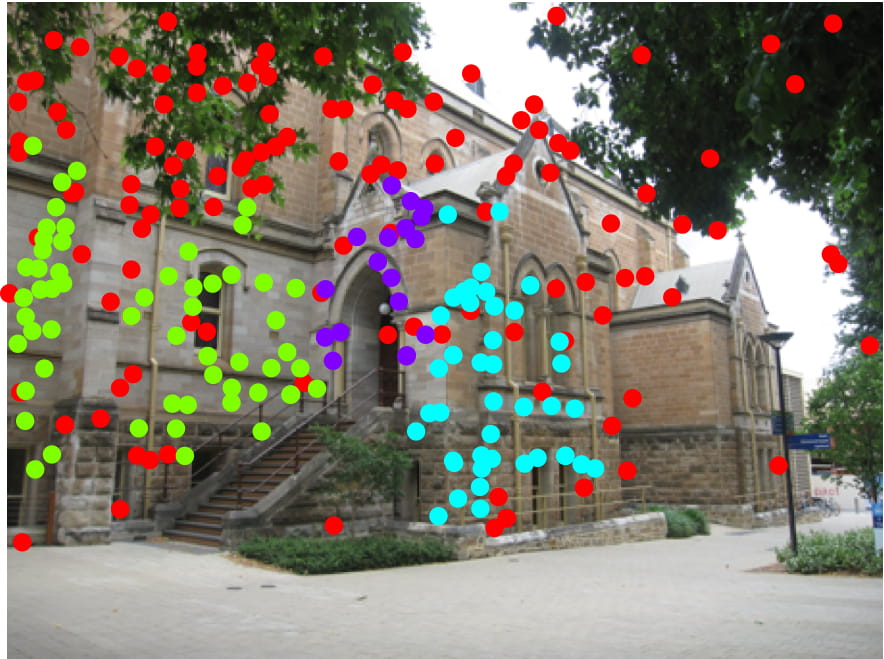}}\\
	\subfigure[Ground-truth]{\includegraphics[width=0.18\linewidth]{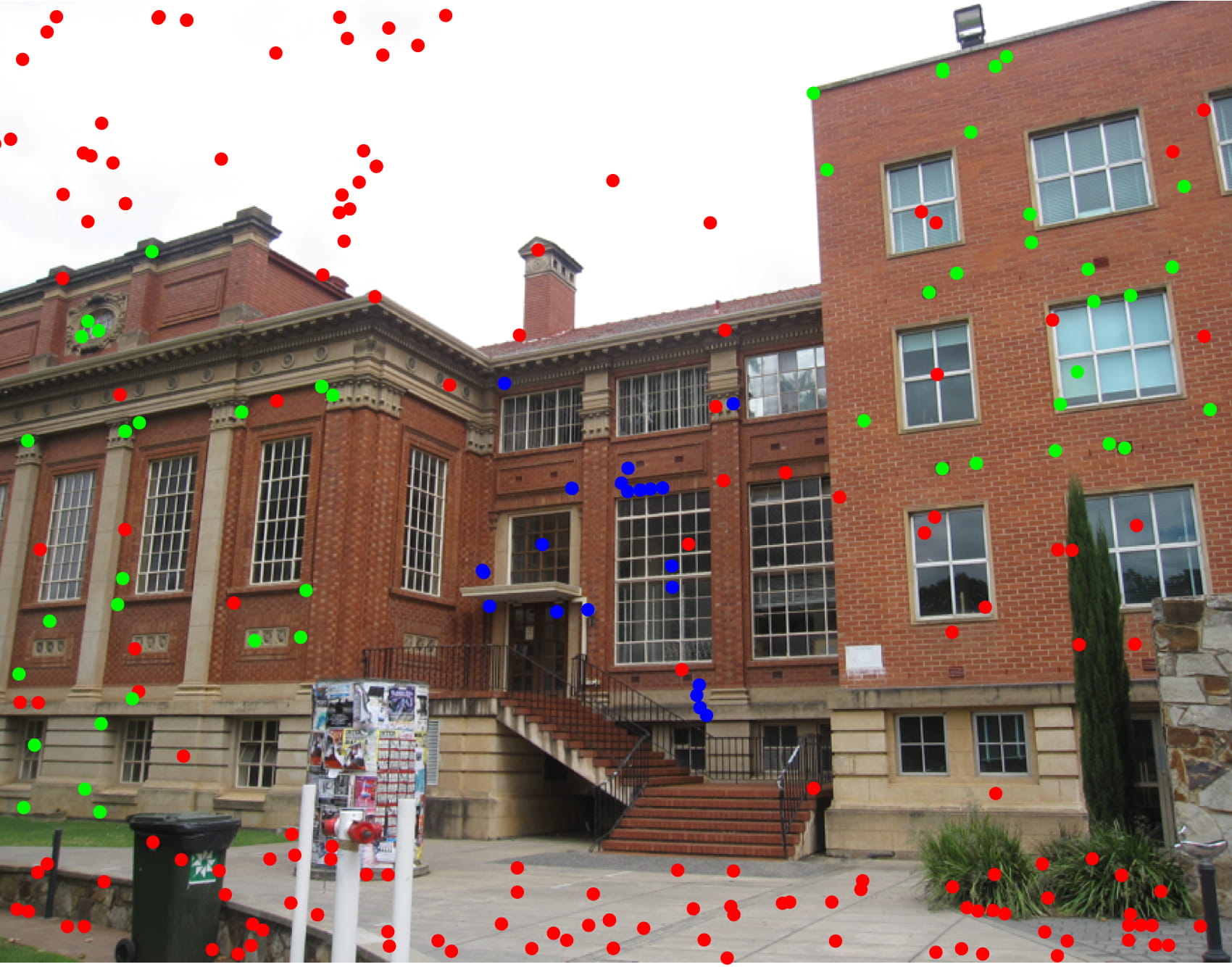}}
	\subfigure[Result of PEARL (10.4\%)]{\includegraphics[width=0.18\linewidth]{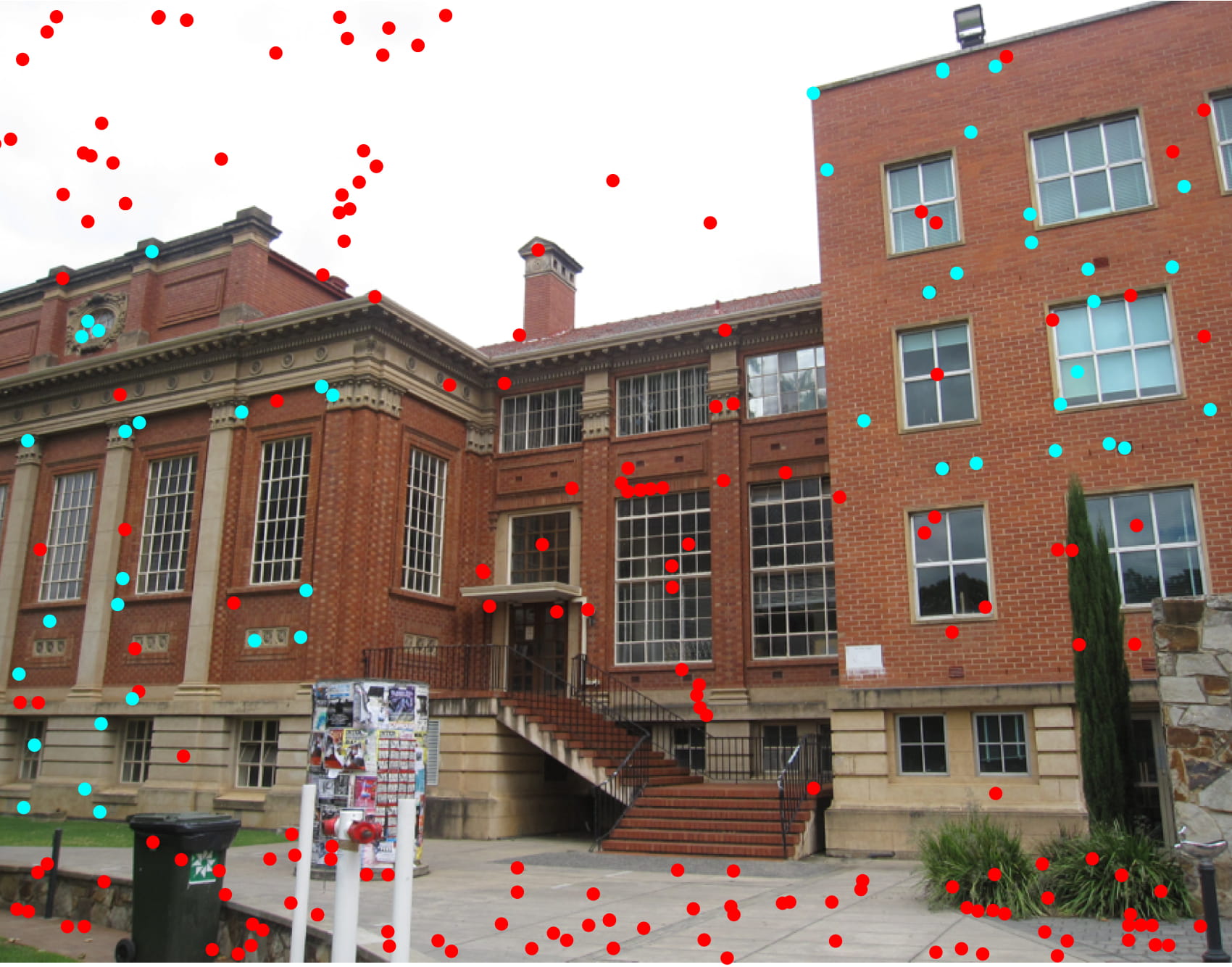}}
	\subfigure[Result of SA-RCM (10.4\%)]{\includegraphics[width=0.18\linewidth]{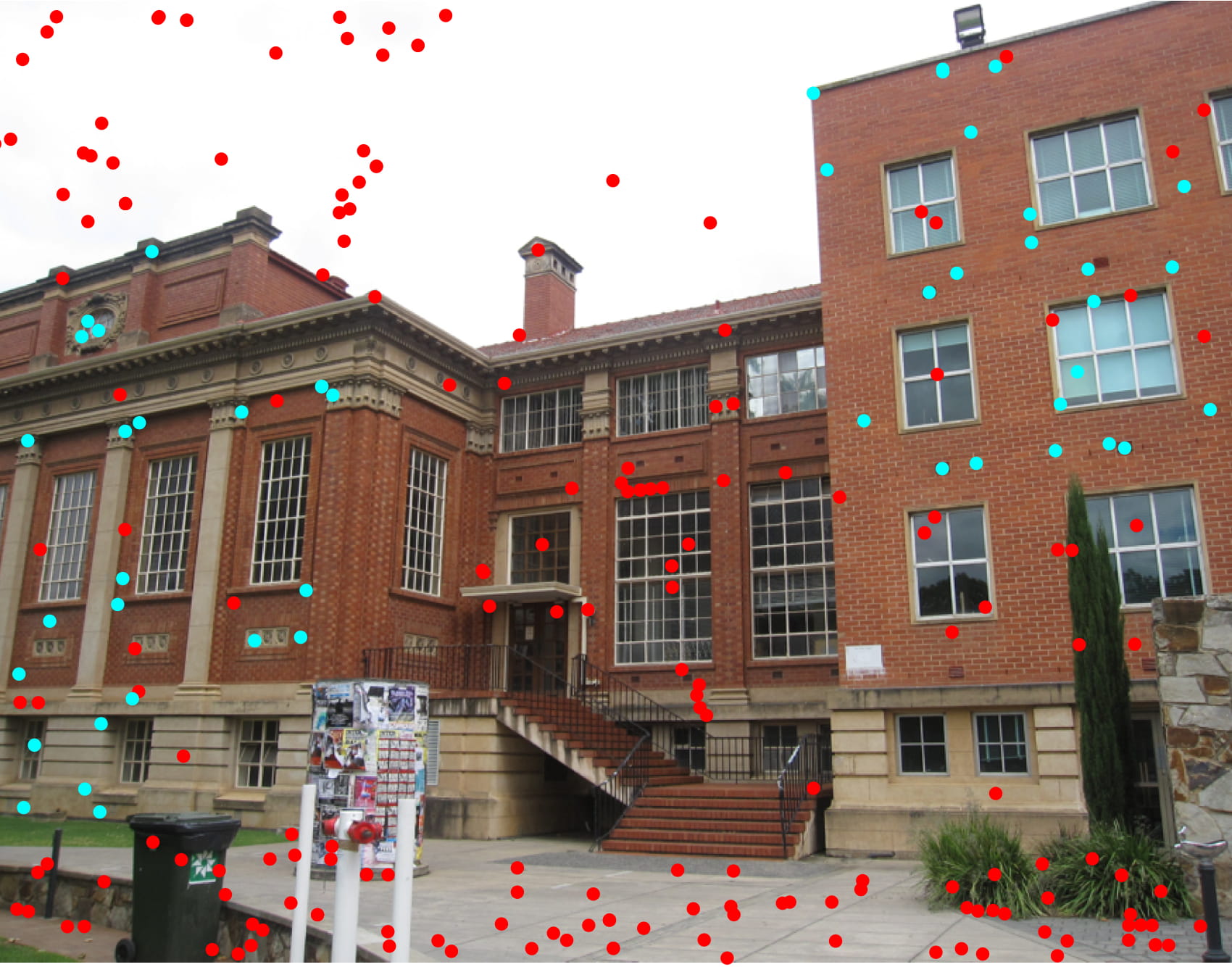}}
	\subfigure[Weak annotations]{\includegraphics[width=0.18\linewidth]{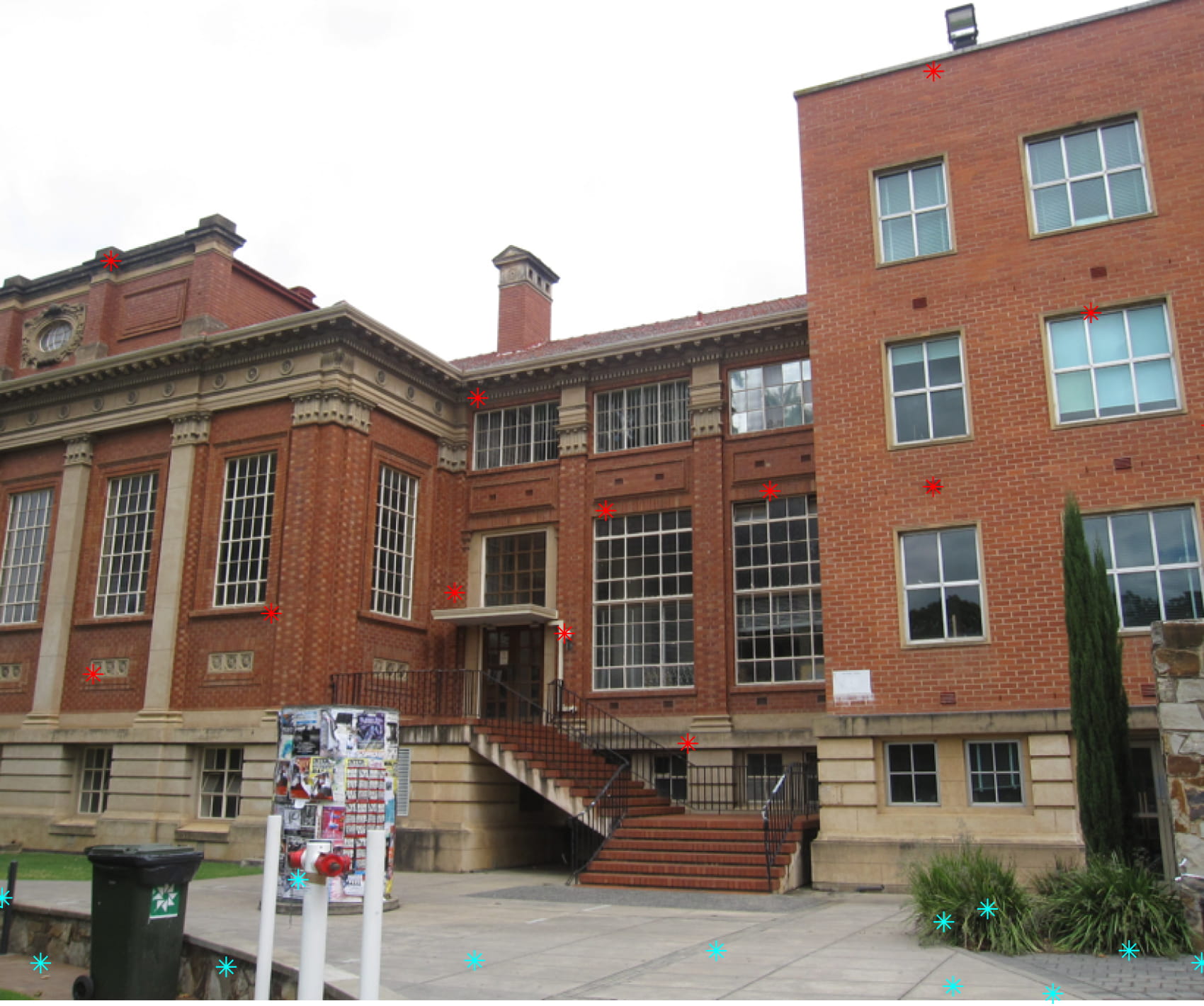}}
	\subfigure[Result of G2MF-WA (2.1\%)]{\includegraphics[width=0.18\linewidth]{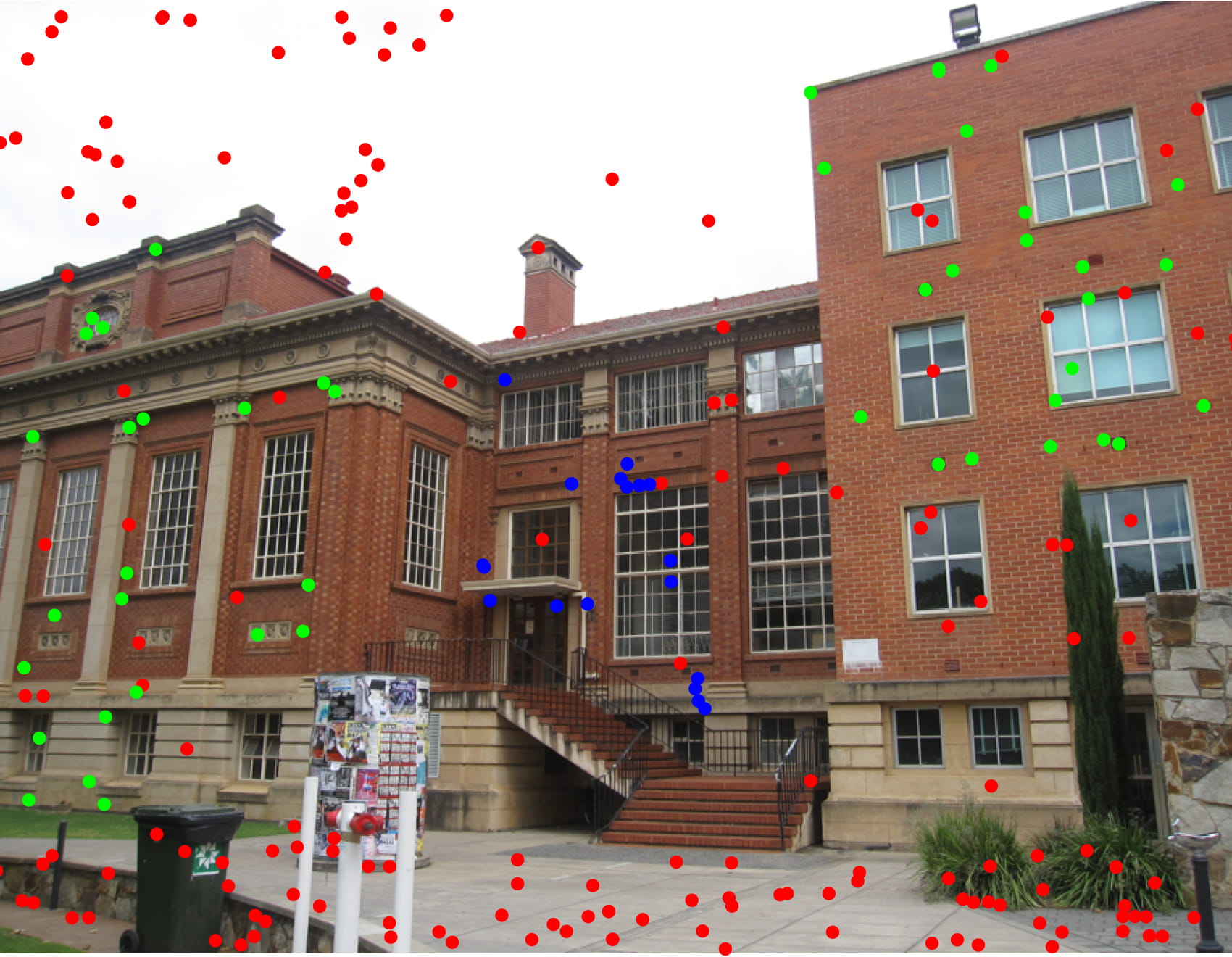}}
	
	\caption{Examples of the multi-homography detection. The first row (\textit{elderhallb}) shows the situation when the number of weak labels in (d) equals the number of ground-truth labels (1st column) \cz{without considering the outlier label (red circle in (a)), i.e., 3 labels in (d) and 4 labels in (a) including the outlier.} The second row (\textit{barrsmith}) shows the situation when the number of weak labels in (i) differs from the ground truth. In (d) and (i), \cz{WA data points are shown}. Best viewed in color.}
	\label{fig:homoexample}
\end{figure*}

\cz{Unlike \cite{pham2014random} that only used part benchmark data, we evaluate full benchmarks in both applications.
	Although the difference between the number of ground-truth labels and the number of WA labels is supposed to affect the final labeling result, our method can still detect the homography/fundamental matrices accurately (e.g., second row in Fig. \ref{fig:homoexample}). One potential reason could be the robustness of the $\alpha$-expansion algorithm to the initial estimate \cite{Isack2012}. Also, although the WA data imposes priors on the edge probabilities between vertices in the sample graph, which allows the algorithm to generate proposals close to the intent of the annotator, the randomness of proposal generation is still included to keep the proposal diversity. This could be another possible factor that contributes to the above finding.}


\subsection{Application 3: planar augmented reality application}
\begin{figure}[tb]
	\centering
	\includegraphics[width=1\linewidth]{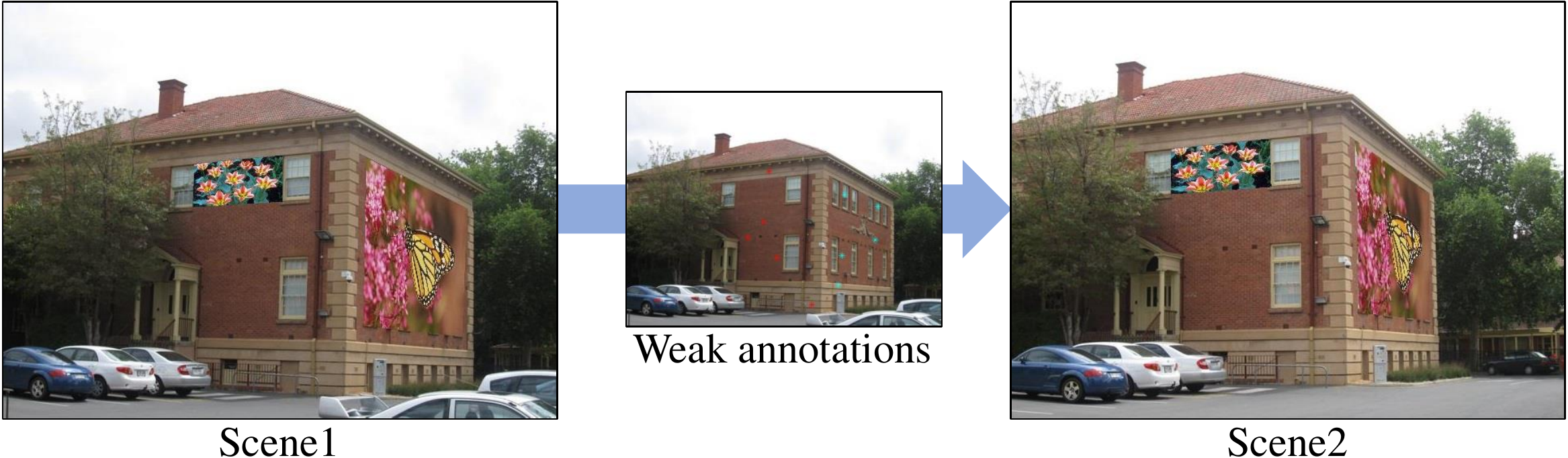}
	\caption{\czcz{Planar augmented reality application based on multi-homography detection with weak annotations. In scene1, two synthetic images are photo-realistically inserted into two manually specified regions, respectively. To adaptively transform the inserted images when scene1 changes to scene2, G2MF-WA requires several weak annotations which can be achieved interactively based on matched keypoint pairs.}}
	\label{fig:AR}
\end{figure}

\czcz{We show a real-world application of planar augmented reality in Fig. \ref{fig:AR}, which is required to insert multiple prepared images to the planar structures in the scene. Here, the visual satisfaction closely relates to the detection accuracy of the planar surfaces in a scene. Our algorithm is designed to improve the accuracy with the help of additional weak annotations. The WA data is obtained by the users in an interactive style.}

\section{Conclusion}
In this paper, we have presented a multi-model fitting method with the assistance of weakly annotated data. The main contribution is to take advantage of the prior knowledge brought by the weakly annotated data, and incorporate it into the calculation of edge probabilities in the proposal sampling graph for effective model proposal generation and further labeling. Extensive experiments demonstrate that our method mostly outperforms the state-of-the-art methods in terms of both accuracy and runtime. 

Despite the effectiveness of our method, it still has a few limitations. Since the model proposals are explored heuristically, the fitting performance is likely to depend on random seeding, and the segmentation error might grow while the algorithm gets stuck in local optimum. One potential way to solve these issues is to increase the number of iterations of the simulated annealing algorithm. As the future work, we would like to design an interactive annotation interface and embed it within the proposed framework. \czcz{Also, we plan to improve the usability of our algorithm by further reducing the user effort.}

\CvmAck{Chao Zhang is supported in part by JSPS KAKENHI Grant JP18K17823. Xuequan Lu is supported in part by Deakin CY01-251301-F003-PJ03906-PG00447.
}

\bibliographystyle{CVM}
{\bibliography{reference}

\Author{a1}{Chao Zhang}
{received his Ph.D. at Iwate university (Japan) in March 2017. He is now a full-time assistant professor at faculty of engineering, university of Fukui (Japan). His research interests include computer vision and graphics, mainly focused on feature matching and vision-based optimization problems. He is a member of the IEEE computer society, IEEE signal processing society, ACM and IEICE.}

\Author{a2}{Xuequan Lu}
{is a Lecturer (Assistant Professor) at Deakin University, Australia. He spent more than two years as a Research Fellow in Singapore. Prior to that, he earned his Ph.D at Zhejiang University (China) in June 2016. His research interests mainly fall into the category of visual computing, for example, geometry modeling, processing and analysis, animation/simulation, 2D data processing and analysis. More information can be found at \url{http://www.xuequanlu.com}}

\Author{a3}{Katsuya Hotta}
{received the B.E. degree in 2017 and is now pursuing Ph.D. degree at University of Fukui, Fukui city, Japan. His current research focuses primarily on the computer vision, mainly for subspace clustering and visual tracking.}

\Author{a4}{Xi Yang}
{is currently a project assistant professor in Graduate School of Information Science and Technology at The University of Tokyo. He received the BE degree in College of Information Engineering from Northwest A\&F University in 2012. He received the ME, DE degrees in Graduate School of Engineering at Iwate University. His research interests include geometric processing, visualization and deep learning.}

}
\end{document}